\documentclass[lettersize,journal]{IEEEtran}
\usepackage{amsmath,amsfonts}
\usepackage{amssymb}
\usepackage{algorithmic}
\usepackage{algorithm}
\usepackage{array}
\usepackage[caption=false,font=footnotesize,labelfont=rm,textfont=rm]{subfig}
\usepackage{textcomp}
\usepackage{stfloats}
\usepackage{url}
\usepackage{verbatim}
\usepackage{graphicx}
\usepackage{cite}
\usepackage{booktabs}
\usepackage{multirow}
\usepackage{balance}
\usepackage{marvosym}
\usepackage{etoolbox}
\usepackage{xcolor}
\usepackage{pifont}
\usepackage[inkscapelatex=false]{svg}
\usepackage{caption}
\usepackage[breaklinks,colorlinks,linkcolor=red,citecolor=blue]{hyperref}

\usepackage{orcidlink}

\AtEndPreamble{
    \usepackage[capitalize]{cleveref}
    \crefname{section}{Sec.}{Secs.}
    \Crefname{section}{Section}{Sections}
    \Crefname{table}{Table}{Tables}
    \crefname{table}{Tab.}{Tabs.}
}

\definecolor{forestgreen}{HTML}{228B22}
\definecolor{lightgreen}{RGB}{169,209,142}
\definecolor{trajgreen}{RGB}{55,120,34}

\makeatletter
\DeclareRobustCommand\onedot{\futurelet\@let@token\@onedot}
\def\@onedot{\ifx\@let@token.\else.\null\fi\xspace}

\patchcmd{\subsubsection}{\itshape}{\bfseries\upshape}{}{}  

\def\@IEEEBIOskipN{1.5\baselineskip}
\patchcmd{\IEEEbiography}
  {\vskip \@IEEEBIOskipN plus 1fil minus 0\baselineskip}
  {\vskip \@IEEEBIOskipN}
  {}{}
\makeatother

\usepackage{xspace}

\newcommand{\negpoint}{indirect\xspace}
\newcommand{\pospoint}{spatially explicit\xspace}

\newcommand{\frkfull}{spatial-inspection-guided ACtive Exploration\xspace}
\newcommand{\frkabbr}{ACE\xspace}

\newcommand{\hpone}{evidence-grounded perception\xspace}

\newcommand{\hponegnr}{Evidence-grounded perception\xspace}

\newcommand{\hptwo}{exposure-informed movement\xspace}

\newcommand{\hptwognr}{Exposure-informed movement\xspace}

\newcommand{\cponefull}{Discriminative Probing\xspace}
\newcommand{\cponeabbr}{DP\xspace}

\newcommand{\cponeloc}{granular localization\xspace}
\newcommand{\cponelocfull}{Granular Localization\xspace}
\newcommand{\cponever}{focused verification\xspace}
\newcommand{\cponeverfull}{Focused Verification\xspace}

\newcommand{\cptwo}{prospective prioritization\xspace}
\newcommand{\cptwofull}{Prospective Prioritization\xspace}
\newcommand{\cptwoabbr}{PP\xspace}

\newcommand{\cpthree}{retrospective suppression\xspace}
\newcommand{\cpthreefull}{Retrospective Suppression\xspace}
\newcommand{\cpthreeabbr}{RS\xspace}

\newcommand{\effone}{inferential validity\xspace}
\newcommand{\efftwo}{directional judgment\xspace}

\newcommand{\titlename}{Active Spatial Inspection for Effective and Efficient Embodied Exploration}

\begin{document}

\title{\titlename}

\author{Wenbin Wang\,\orcidlink{0000-0002-4394-0145},~\IEEEmembership{Member,~IEEE}, 
        Xiang Bai\,\orcidlink{0000-0002-3449-5940},~\IEEEmembership{Fellow,~IEEE},
        Yizhao Wang\,\orcidlink{0000-0002-8990-6663},
        Hang Sun\,\orcidlink{0000-0002-3460-9354},
        Dong Ren\,\orcidlink{0009-0001-1048-5393},
        
        Jie Qin\,\orcidlink{0000-0002-0306-534X},~\IEEEmembership{Member,~IEEE},
        Qingquan Li\,\orcidlink{0000-0002-2438-6046}, 
        Bing Wang\,\orcidlink{0000-0003-0977-0426},~\IEEEmembership{Member,~IEEE}

\thanks{This work is supported in part by the Natural Science Foundation of China (42671592, 42301520, 62506207), the Research Grants Council of Hong Kong (25206524, 15212925), the Scientific and Technological Research Project of the Department of Education of Hubei Province (D20251204), and the Smart Cities Research Institute (P0051028). (\emph{Corresponding author: Bing Wang.}) }

\thanks{Wenbin Wang is with The Hong Kong Polytechnic University, and with China Three Gorges University (e-mail: nkwangwenbin@gmail.com). }

\thanks{Xiang Bai is with Huazhong University of Science and Technology (e-mail: xbai@hust.edu.cn).}

\thanks{Yizhao Wang and Bing Wang are with The Hong Kong Polytechnic University (e-mail: yizhao.wang@polyu.edu.hk, bingwang@polyu.edu.hk). }

\thanks{Hang Sun and Dong Ren are with China Three Gorges University (e-mail: sunhang@ctgu.edu.cn, dren@ctgu.edu.cn).}

\thanks{Qingquan Li is with Shenzhen University (e-mail: liqq@szu.edu.cn).}

\thanks{Jie Qin is with Nanjing University of Aeronautics and Astronautics (e-mail: jie.qin@nuaa.edu.cn).}

}

\maketitle

\begin{abstract}
Achieving high task success and efficiency remains a central pursuit in embodied exploration.
Existing frameworks typically guide agent behavior through a spatially coarse and \negpoint assessment of suggestive cues and directions, yet such designs may struggle to judge cue sufficiency and the need for further inspection, leading to early termination or excessive continuation and ultimately reducing task success and efficiency.
This work rethinks embodied exploration from a \pospoint standpoint and introduces the \frkfull (\frkabbr) framework coupling \hpone with \hptwo and establishing a spatially resolved decision paradigm.
\hponegnr strengthens \effone by integrating \cponeloc with \cponever, turning suggestive cues into precise decisive visual support.
\hptwognr promotes \efftwo through \cptwo and \cpthree, selecting promising directions for efficient spatial progress.
\frkabbr alleviates the tension between preventing early termination and avoiding unnecessary continuation.
Extensive experiments demonstrate that \frkabbr achieves 18.0\% higher navigation task success and 10.3\% higher exploration efficiency for question answering than prior state-of-the-art baselines, advancing effective and efficient embodied exploration.

\end{abstract}


\section{Introduction}
\label{sec:intro}

\IEEEPARstart{E}{mbodied} exploration~\cite{yang20253d,gu2024conceptgraphs,jiang2024roboexp} is a fundamental capability for agents in open 3D environments, with broad applications in household assistive robotics~\cite{ehsani2024spoc} and industrial inspection~\cite{pistone2024modelling}. It underpins a wide range of tasks, such as embodied question answering~\cite{majumdar2024openeqa,saxena2024grapheqa}, object navigation~\cite{yin2024sg,batra2020objectnav,yin2025unigoal}, and long-horizon target search~\cite{khanna2024goat,ginting2025enter}.
Throughout exploration, the agent repeatedly faces two closely coupled decisions, i.e., when to terminate and, if not, where to proceed.
Terminating before sufficient task evidence is established can compromise success, whereas continuing after such evidence becomes available incurs unnecessary search.
If further exploration is required, the agent must decide
where to proceed to acquire useful information.
Effective and efficient embodied exploration therefore requires reliable judgments of when to terminate and productive choices of where to proceed.

\begin{figure}[t]
\setlength{\belowcaptionskip}{-0.2cm}
\setlength{\belowcaptionskip}{-0.4cm}
    \centering
    \includegraphics[width=0.9\linewidth]{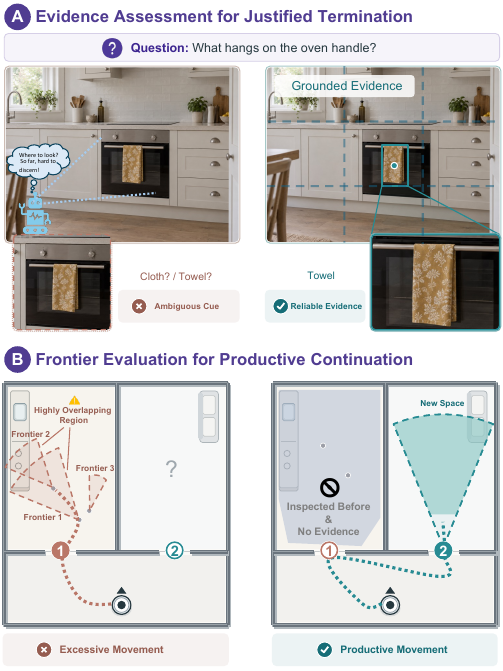}    
    \caption{
    Illustration of the two spatial assessment bottlenecks in embodied exploration and how \frkabbr addresses them.
    (A) Existing methods often reason over coarse visual units, leading to ambiguous termination decisions. \frkabbr instead grounds task-relevant cues into explicit spatial candidates and verifies the localized content to establish reliable evidence for termination.
    (B) Existing methods indirectly indicate whether a space has already been inspected. Different frontiers may lead to substantially overlapping observations and unnecessary movement.
    \frkabbr explicitly relates each frontier to the space it exposes, suppressing exhausted directions while favoring those that can reveal useful information.
    Together, the two aspects illustrate the shift from coarse and indirect assessment to spatially explicit evidence assessment and frontier evaluation.
    }
    \label{fig:evi_properties}
\end{figure}

Recent embodied exploration systems increasingly leverage
VLMs~\cite{bai2025qwen2,achiam2023gpt,wu2025spatial} to support zero-shot perception and decision-making.
Building on VLMs, existing methods have advanced the two
decisions above through richer contextual representations and more explicit decision criteria.
For deciding when to terminate, confidence calibration and
memory-conditioned reasoning assess whether available observations support an answer~\cite{ren2024explore,saxena2024grapheqa,zhai2025memoryeqa}, while re-perception and goal verification re-examine candidate targets before task completion~\cite{yin2024sg,yin2025unigoal}.
For deciding where to proceed, exploration policies combine semantic relevance with frontier potential~\cite{yokoyama2023vlfm,wang2026scope},
while visitation feedback, reflection, and semantic prediction updates refine subsequent choices~\cite{metanav2026,chen2026hgr}.
Despite these advances, current methods still exhibit limitations in evidence assessment and frontier evaluation. These limitations can undermine both termination reliability and exploration efficiency, motivating a closer investigation into their underlying causes.

\IEEEpubidadjcol

A key bottleneck in termination lies in how task-relevant visual cues are exposed for evidence assessment. Existing methods reason over retrieved snapshots~\cite{yang20253d} or detected object instances~\cite{yin2024sg,yin2025unigoal}, with re-perception or goal verification providing additional checks before task completion. These mechanisms improve decision reliability. However, as shown in \cref{fig:evi_properties}(A), a task-relevant snapshot may contain a small region with decisive evidence that is hard to discern through whole-view reasoning, while a visible target missed by the detector provides no object candidate that can be selected and further inspected. Consequently, recognizing a relevant view does not necessarily reveal sufficient evidence, and a missing detection does not mean that useful cues are absent. When assessment relies only on coarse views or detector-defined objects, agents may terminate on ambiguous cues or continue searching even when decisive evidence is already visible.

A parallel bottleneck in continuation lies in distinguishing spatial visitation from actual visual inspection (\cref{fig:evi_properties}(B)). Existing methods use visitation history to reduce redundant movement~\cite{metanav2026}, while semantic- or potential-based reasoning helps rank candidate frontiers~\cite{wang2026scope,chen2026hgr}. These methods provide useful guidance, but they only approximate whether a space has already been sufficiently inspected. 
A frontier is defined not only by its position but also by the direction it faces. Nearby frontiers may expose different regions, while spatially separated frontiers may lead to overlapping views. Semantic relevance likewise does not by itself indicate whether the corresponding space has already been inspected. For judging redundant exploration, the more direct question is whether the space exposed by a candidate frontier has already been inspected. Otherwise, informative directions may be suppressed too early, while already inspected regions may be repeatedly explored, reducing exploration efficiency.

Motivated by these two bottlenecks, we propose the \frkfull (\frkabbr) framework, which couples \textbf{\hpone} with \textbf{\hptwo} to make both evidence assessment and frontier evaluation \pospoint. \frkabbr links task-relevant cues to grounded spatial candidates and candidate frontiers to the space they expose. It verifies whether localized cues provide sufficient support for termination, while evaluating candidate directions by considering both their potential to reveal useful information and whether the associated space has already been sufficiently inspected, thereby supporting justified termination and productive continuation.

To instantiate \hpone, we introduce \textbf{\cponefull}, which strengthens \effone through \cponeloc and \cponever. For answer-oriented tasks, \frkabbr localizes task-relevant regions within retrieved observations and verifies whether the focused visual content provides sufficient support. For goal-oriented tasks, it first examines detected object instances as spatial candidates. When such candidates are missing or unreliable, \frkabbr further introduces detector-independent evidential visual points to recover task-relevant cues directly from raw observations. These points can be spatially grounded, approached, and further verified before termination. Only sufficiently verified candidates can trigger completion, moving task assessment beyond whole-view judgments and detector-defined objects for more reliable and timely termination.

To instantiate \hptwo, we develop two complementary mechanisms. \textbf{\cptwofull} first establishes a prospective basis for frontier evaluation by combining visible task relevance with each frontier's potential to reveal useful unseen space, ensuring that directions with weak immediate cues are not discarded too early. Building on this prospective estimate, \textbf{\cpthreefull} then determines whether a direction has already been sufficiently inspected, integrating accumulated visual coverage with repeated evidence failures and recent trajectory progress to suppress visually exhausted or low-progress directions. Together, the two mechanisms distinguish frontiers that still warrant further inspection from those whose remaining value has largely been exhausted, thereby improving \efftwo.



The main contributions are summarized as follows:

\begin{itemize}
    \item We present a \pospoint perspective on embodied exploration, emphasizing the spatial localization of task-relevant cues and the inspection status of candidate frontier regions. These insights motivate \frkabbr, which couples spatially explicit evidence assessment with frontier evaluation. This formulation provides a unified basis for justified termination 
    and productive continuation.

    \item We propose \hpone to strengthen \effone. It is realized by \cponefull through \cponeloc and \cponever, allowing task cues 
    to be localized and verified. Particularly, evidential visual points recover visible cues 
    beyond detected objects. This design extends task assessment beyond coarse views and predefined object candidates.

    \item We design \hptwo to improve \efftwo. \cptwofull first identifies informative frontiers by considering visible cues and hidden-space potential. \cpthreefull then integrates visual coverage, evidential feedback, and trajectory 
    to suppress sufficiently inspected directions. This coordinated design preserves worthwhile directions.

    \item Extensive experiments on embodied question answering and single-/multi-goal navigation demonstrate that \frkabbr consistently achieves leading performance, with up to \textbf{18.0\%} higher 
    navigation task success and \textbf{10.3\%} higher exploration efficiency for question answering than the strongest prior baselines. These results provide consistent empirical support for the effectiveness and generality of \frkabbr across diverse embodied tasks.

\end{itemize}

\section{Related Work}
\label{sec:related_work}

\subsection{Evidence Assessment for Termination}
\label{sec:rw_evidence_assessment}

Embodied exploration encompasses tasks in which an agent must acquire information from an unknown environment, including object-goal navigation~\cite{batra2020objectnav,khanna2024goat,zhang2025hoz++}, image-goal navigation~\cite{krantz2022instanceimagenav,lei2025gaussnav}, vision-language navigation~\cite{anderson2018vision,liu2023bird,song2025towards,dai2026history}, embodied question answering~\cite{das2018embodiedqa,majumdar2024openeqa,ren2024explore}, and long-horizon multi-goal navigation~\cite{wani2020multion,khanna2024goat,song2025towards,ehsani2024spoc}. Recent VLMs and Multimodal Large Language Models (MLLMs)~\cite{achiam2023gpt,team2024gemini,li2024llava,bai2025qwen2,wu2025spatial} provide open-vocabulary perception, commonsense reasoning, and instruction-conditioned decision making for these tasks. Existing agents use them to infer answers, targets, or actions from egocentric observations~\cite{majumdar2024openeqa,ren2024explore,cheng2024efficienteqa,gadre2022cow,shah2022lmnav,zhu2025mtu,yin2025unigoal}, supporting open-world exploration beyond conventional closed-set policies~\cite{zhou2025reinforced,wijmans2019embodied}.
Memory further expands the information available for task assessment by organizing temporally dispersed observations. Spatial and semantic memories represent explored environments as metric maps~\cite{shafiullah2022clip,huang23vlmaps,yamazaki2024open}, open-vocabulary 3D representations~\cite{jatavallabhula2023conceptfusion,kerr2023lerf,huang2025multimodal}, or scene graphs~\cite{armeni20193d,rosinol2021kimera,gu2024conceptgraphs,werby2024hovsg}. Image-centric memories make past observations accessible through retrieval, re-observation, or long-context visual reasoning~\cite{yang20253d,zhai2025memoryeqa,li2026himm,lu2026gsmem,fan2025embodiedvideoagent,hu2026astranav}.

These approaches substantially improve the availability of task-relevant information, but task assessment is commonly performed over relatively coarse visual units. As a result, suggestive cues may remain insufficiently resolved. Decisive local content can be obscured within a full view, while visible targets missed by a detector may never become candidates for further assessment. In contrast, \frkabbr resolves task-relevant cues into localized visual candidates and verifies whether they provide sufficient support before termination.

\subsection{Frontier Evaluation for Continuation}
\label{sec:rw_frontier_valuation}

Frontiers, defined as boundaries between explored free and unobserved regions, provide a compact interface between spatial mapping and planning. Conventional frontier policies prioritize reachable candidates according to geometric factors such as distance, accessibility, and expected coverage. With foundation models, frontier evaluation has become increasingly task-conditioned. Recent methods use language-grounded value maps and task relevance~\cite{yokoyama2023vlfm,ren2024explore}, frontier-facing snapshots~\cite{yang20253d}, topological ghost nodes~\cite{cui2024frontier}, spatio-temporal semantic propagation~\cite{wang2026scope}, or revisable semantic hypotheses~\cite{chen2026hgr} to estimate the utility of unexplored regions. These developments turn frontiers from geometric boundaries into semantic interfaces for prospective exploration.

Long-horizon agents also exploit prior exploration experience to update subsequent decisions. Such experience can be recorded within the current trajectory or accumulated across episodes~\cite{wang2024karma,zhang2025reexplore,metanav2026,dai2026history}, represented as revisable hypotheses~\cite{chen2026hgr}, or extended through imagined future observations and states~\cite{yang2025mindjourney,wang2026imaginenav++,hu2025imaginative}. These methods demonstrate the value of both prospective reasoning and retrospective feedback. However, the inspection status of frontier-associated space is often inferred indirectly from semantic, geometric, or historical cues. In contrast, \frkabbr incorporates explicit visual coverage into frontier evaluation, preserving promising directions while suppressing exhausted ones.

\subsection{Active Information Seeking in Embodied Exploration}
\label{sec:rw_active_evidence_seeking}

Active information seeking has long been studied as a means of acquiring task-relevant information beyond passive sensory accumulation. Cognitive studies show that goal-oriented attention prioritizes diagnostic stimuli~\cite{corbetta2002control}, while active sampling and curiosity adapt information acquisition according to uncertainty and expected information value~\cite{gottlieb2013information,gottlieb2018active_sampling,petitet2021active_sampling,balsdon2020confidence}. Related principles have been adopted in embodied AI through active perception~\cite{bajcsy2018revisiting}, curiosity-driven exploration~\cite{pathak2017curiosity}, and uncertainty-aware navigation~\cite{ruan2026brain}. However, existing approaches commonly instantiate active information seeking within individual decision stages, such as frontier ranking, uncertainty estimation, reflection, or memory retrieval. Their resulting assessments are therefore typically tied to the representation used at that stage, rather than explicitly resolving where decisive support lies or where further observation remains worthwhile. \frkabbr instead adopts a spatially resolved view of active information seeking across both termination and continuation, connecting task-relevant cues to localized visual support and candidate directions to the space available for further observation.

\section{Preliminaries}
\label{sec:preliminary}

Embodied exploration is a sequential decision process in which an agent interacts with an initially unknown 3D environment to complete a task. Let $q$ denote the task specification, which can be an answer-oriented question or a goal-oriented navigation instruction. At step $t$, the agent pose is $\mathbf{x}_t=(\mathbf{p}_t,\psi_t)$, where $\mathbf{p}_t$ is the position and $\psi_t$ is the orientation, represented by the yaw angle in the horizontal plane. The agent captures a set of egocentric RGB-D observations $\mathcal{I}_t=\{I_t^1,I_t^2,\ldots,I_t^{N_t}\}$, and executes an action $a_t$ based on $q$, $\mathbf{x}_t$, $\mathcal{I}_t$, and the accumulated knowledge state $\mathcal{K}_t$.

In this work, the accumulated knowledge state $\mathcal{K}_t$ consists of two types of memories. The first type follows common designs in existing embodied exploration systems and records directly observable environmental information, including the snapshot memory $\mathcal{S}_t$, the frontier set $\mathcal{F}_t$, and the object memory $\mathcal{O}_t$. These components summarize what objects have been observed, which egocentric views can represent the explored scene, and where unexplored regions remain accessible, providing the basic visual and spatial substrate for identifying and verifying task-relevant evidence: 

\texttt{Snapshot Memory} $\mathcal{S}_t$ provides a compact representation of the egocentric observation history. Following 3D-Mem~\cite{yang20253d}, \frkabbr condenses the raw observation stream $\mathcal{I}_t$ into a set of representative key snapshots $\{S_k\}$. These snapshots are updated online by spatially grouping observed objects and selecting representative views that collectively cover the explored objects. This compact snapshot memory supports subsequent evidence probing and long-horizon exploration planning.

\texttt{Frontier Set} $\mathcal{F}_t$
contains frontiers that are connected boundaries between the observed free space and the unknown space. Concretely, RGB-D observations are fused into a Truncated Signed Distance Function (TSDF)-based voxel volume $\mathcal{V}_t$, whose voxels are partitioned into free, occupied, and unknown subsets:
\begin{equation}
\mathcal{V}_t=\mathcal{V}_t^{\mathrm{free}}\cup
\mathcal{V}_t^{\mathrm{occ}}\cup\mathcal{V}_t^{\mathrm{unk}} .
\end{equation}
We denote frontier set by $\mathcal{F}_t=\{F_t^i\}_{i=1}^{M_t}$, where $M_t$ denotes the number of frontiers, and $F_t^i$ satisfies
\begin{equation}
F_t^i\subset\mathcal{V}_t^{\mathrm{free}},\quad
    \forall \mathbf{v}\in F_t^i,\exists \mathbf{v}'\in\mathcal{N}(\mathbf{v}),
    \mathbf{v}'\in\mathcal{V}_t^{\mathrm{unk}}.
    \label{eq:frontier_def}
\end{equation}
$\mathcal{N}(\mathbf{v})$ is the spatial neighborhood of voxel $\mathbf{v}$. Each frontier $F_t^i$ has a centroid $\mathbf{f}_t^i$, an outward direction $\mathbf{d}_t^i$, a spatial mask $m(F_t^i)$ that reveals which regions will be explored, and a frontier-facing visual observation $I(F_t^i)$.

\texttt{Object Memory} $\mathcal{O}_t$ is obtained by merging detected objects from local observations and associated snapshots into persistent 3D object instances. We denote it as
\begin{equation}
    \mathcal{O}_t=\{o_j=(c_j,\mathbf{z}_j,\sigma_j)\}_{j=1}^{L_t},
\end{equation}
where $L_t$ is the number of objects, $c_j$ is the semantic label, $\mathbf{z}_j$ is the 3D location, and $\sigma_j$ is the detection confidence.

The second type is introduced by \frkabbr. Specifically, the  \texttt{History Memory} $\mathcal{H}_t$ stores the past frontier-driven motion outcomes. The \texttt{Visual Coverage Memory} $\mathcal{C}_t$ records the spatial regions that have been visually inspected from previous viewpoints. The 
\texttt{Frontier Status Memory} $\mathcal{R}_t$ maintains the estimated place type and evidential staleness of previously selected frontier directions. We introduce these memories in the following section.


\section{Methodology}
\label{sec:method}

\begin{figure*}[t]
\setlength{\belowcaptionskip}{-0.5cm}
    \centering
    \includegraphics[width=\linewidth]{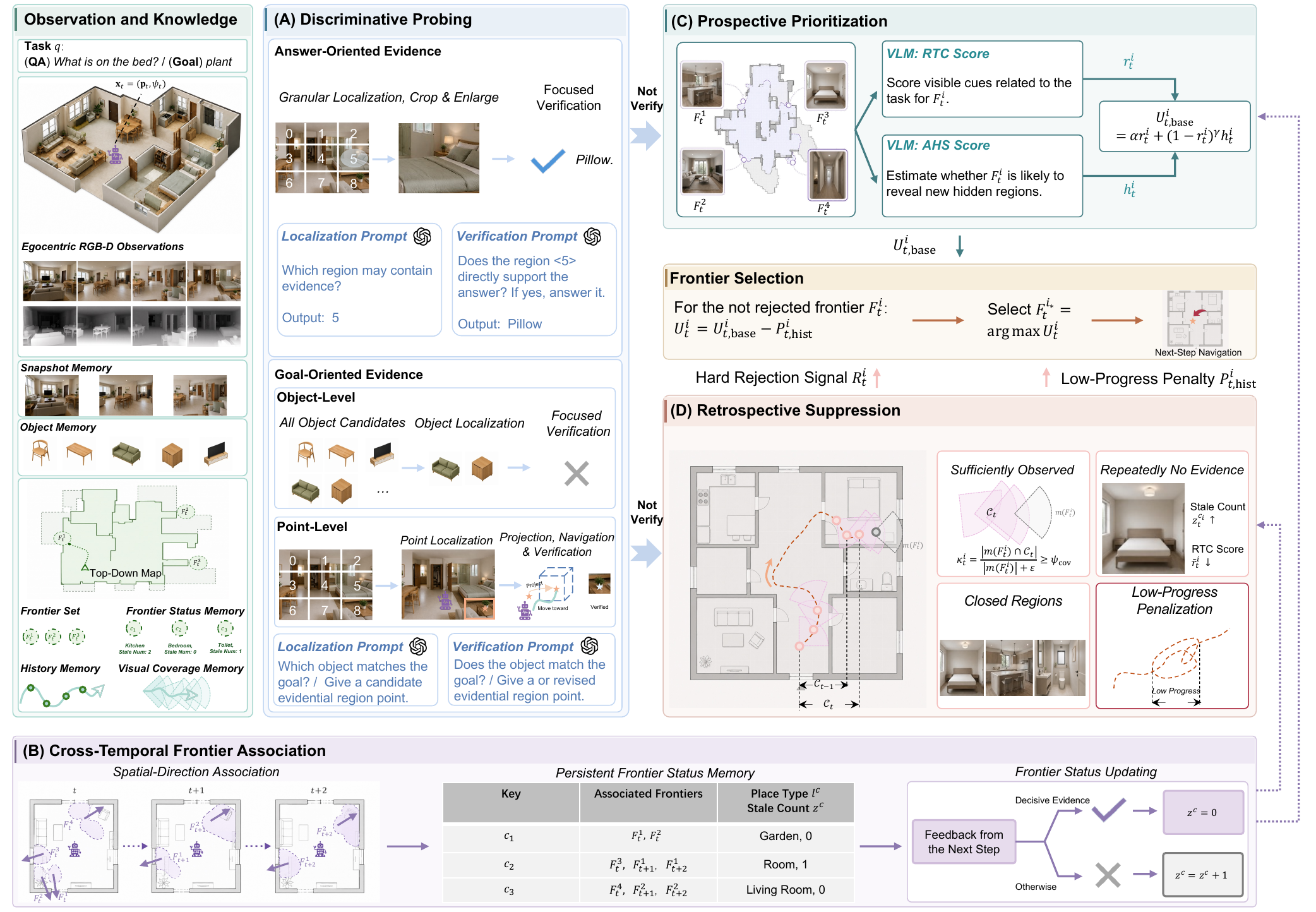}
    \caption{Overview of the proposed \frkabbr framework. Given the current observations and accumulated knowledge, (A) \cponefull localizes and verifies evidence for termination. (B) Cross-Temporal Frontier Association links transient frontier instances across decision steps, providing a persistent association basis for frontier evaluation and regulation. If decisive evidence is unavailable, (C) \cptwofull estimates frontier utility from visible task cues and hidden-space potential. (D) \cpthreefull rejects exhausted directions and penalizes low-progress exploration using accumulated feedback. The remaining frontier with the highest regulated utility is selected for continued exploration.}
    \label{fig:method_overview}
\end{figure*}

As illustrated in \cref{fig:method_overview}, \frkabbr is designed to support spatially resolved decision making in embodied exploration by coupling \hpone with \hptwo. At each step $t$, the agent incorporates egocentric observations into the accumulated knowledge state $\mathcal{K}_t$. \frkabbr then makes sequential decisions based on $\mathcal{K}_t$ with three core mechanisms supported by an auxiliary cross-temporal frontier association mechanism. \cponefull (\cref{sec:method_evidence}) localizes and verifies task-relevant visual support for termination. Cross-Temporal Frontier Association (\cref{sec:frontier_association}) links transient frontiers across decision steps. Using these persistent associations, \cptwofull (\cref{sec:method_frontier}) favors frontiers with strong potential to expose useful space, while \cpthreefull (\cref{sec:method_exhaustion}) suppresses exhausted directions and penalizes low-progress ones according to accumulated visual coverage and exploration feedback. The procedure of \frkabbr is summarized in \cref{alg:ace}.

\begin{algorithm}[ht]
\caption{Procedure of \frkabbr}
\label{alg:ace}
\begin{algorithmic}[1]
\REQUIRE Task specification $q$, initial pose $\mathbf{x}_0$, initial knowledge state $\mathcal{K}_0$
\ENSURE Task result and executed trajectory
\STATE Initialize the pending frontier key $c_{\star}\leftarrow\varnothing$
\FOR{$t=0,1,\ldots,T-1$}
    \STATE Capture $\mathcal{I}_t$ at $\mathbf{x}_t$, and detect objects
    \STATE Update $\mathcal{S}_t=\mathcal{S}_{t-1}\cup$\{new key snapshots\}, $\mathcal{O}_t=\mathcal{O}_{t-1}\cup$\{new detected objects\}, $\mathcal{F}_t=\mathcal{F}_{t-1}\cup$\{new frontiers\} 
    \STATE Update $\mathcal{C}_t$ from $\mathcal{C}_{t-1}$ with Eq.~\eqref{eq:visual_coverage}
    \STATE Localize $\hat{\mathcal{E}}_t^\mu$ by Eq.~\eqref{eq:evidence_generation}, enlarge or navigate toward it, and verify it to obtain $\mathcal{E}_t^\mu$ by Eq.~\eqref{eq:verified_evidence_set}
    \IF{$\mathcal{E}_t^\mu \neq \varnothing$}
        \STATE \textbf{return} task result
    \ENDIF
    
    \IF{$c_{\star}\neq\varnothing$}
        \STATE Update $\mathcal{R}_t=\mathcal{R}_{t-1}\setminus \{c_{\star}\mapsto (\ell^{c_{\star}},z^{c_\star})\} \cup \{c_{\star}\mapsto (\ell^{c_{\star}},(z^{c_\star}+1)\,\, \mathrm{or}\,\, 0)\}$ using Eq.~\eqref{eq:stale_update}
        \STATE Set $c_{\star}\leftarrow\varnothing$
    \ENDIF

        \FOR{each frontier $F_t^i\in\mathcal{F}_t$}
            \STATE Compute $c_t^i$ by Eq.~\eqref{eq:key_discr}, and then retrieve $z_t^{c_i}$ and $\ell_t^{c_i}$ from $\mathcal{R}_t$. 
            \STATE Compute $r_t^i$ and $h_t^i$ by Eqs.~\eqref{eq:reactive_score} and \eqref{eq:hidden_score}
            \STATE Compute $U_{t,\mathrm{base}}^i$ by Eq.~\eqref{eq:base_frontier_score}
            \STATE Compute $R_t^i$ by Eq.~\eqref{eq:visual_reject}
            \STATE Compute $P_{t,\mathrm{hist}}^i$ by Eqs.~\eqref{eq:history_penalty}--\eqref{eq:low_progress_penalty}
            \STATE Compute regulated utility $U_t^i$ by Eq.~\eqref{eq:final_frontier_score}
        \ENDFOR
        \STATE Select $F_t^{i_\star}=\arg\max_{F_t^i\in\mathcal{F}_t}U_t^i$ and set $c_\star=c_t^{i_\star}$
        \STATE Execute frontier-guided action toward $F_t^{i_\star}$
        \STATE Set $\mathcal{H}_{t+1}=\mathcal{H}_t\cup\{(\mathbf{p}_t^s,\mathbf{p}_t^e)\}$
\ENDFOR
\end{algorithmic}
\end{algorithm}

\subsection{\cponefull}
\label{sec:method_evidence}

Accurate and timely termination requires task-relevant visual support to be resolved at an appropriate spatial scale. Holistic VLM reasoning over entire egocentric images can overlook small decisive regions, while detector-defined candidates may miss visible cues altogether. To address these limitations, \cponefull performs task-adaptive \cponeloc over regional, object-level, and detector-independent evidential visual-point candidates, followed by \cponever to confirm that the localized candidate is valid for the current task.

\subsubsection*{\cponelocfull}
We represent an evidence proposal as $\widehat{e}=(s,r,\mathbf{y})$, where $s$ is the source view or snapshot, $r$ denotes a localized region when applicable, and $\mathbf{y}$ is the task-specific evidence content.  For each evidence type $\mu$, \frkabbr first generates candidate evidence proposals from the corresponding source set:
\begin{equation}
    \widehat{\mathcal{E}}_t^\mu
    =G_{\mathrm{loc}}^\mu(q,\mathcal{X}_t^\mu)=\{\widehat{e}_k^\mu\},
    \quad
    \mathcal{X}_t^\mu\subseteq
    \{\mathcal{I}_t,\mathcal{S}_t,\mathcal{O}_t\},
    \label{eq:evidence_generation}
\end{equation}
where $\mu\in\{\mathrm{ans},\mathrm{obj},\mathrm{vp}\}$ denotes answer, object, and visual-point evidence.
The localization function $G_{\mathrm{loc}}^\mu$ is instantiated according to the granularity required by each evidence type:
\begin{equation}
    (G_{\mathrm{loc}}^\mu,\mathcal{X}_t^\mu)=
    \begin{cases}
    (G_{\mathrm{loc}}^{\mathrm{ans}},\mathcal{S}_t), & \mu=\mathrm{ans},\\
    (G_{\mathrm{loc}}^{\mathrm{obj}},\{\mathcal{S}_t,\mathcal{O}_t\}), & \mu=\mathrm{obj},\\
    (G_{\mathrm{loc}}^{\mathrm{vp}},\mathcal{I}_t), & \mu=\mathrm{vp}.
    \end{cases}
    \label{eq:evidence_instance}
\end{equation}

\subsubsection*{\cponeverfull}
Candidate proposals are not accepted upon generation. Instead, \frkabbr directs the agent toward each candidate to acquire more informative observations, through which the candidate is verified before being promoted to task-sufficient evidence. \frkabbr retains only proposals that pass \cponever by function $G_{\mathrm{ver}}^\mu$:
\begin{equation}
    \mathcal{E}_t^\mu = G_{\mathrm{ver}}^\mu(q, \widehat{\mathcal{E}}_t^\mu)
    =\{e_k^\mu\},
    \label{eq:verified_evidence_set}
\end{equation}
where $\mathcal{E}_t^\mu$ denotes the set of the accepted evidence $e_k^\mu$, $G_{\mathrm{ver}}^\mu$ is instantiated to $G_{\mathrm{ver}}^{\mathrm{ans}}$, $G_{\mathrm{ver}}^{\mathrm{obj}}$, and $G_{\mathrm{ver}}^{\mathrm{vp}}$, respectively.

The concrete form of the evidence proposal $\widehat{e}$ or the evidence $e$ depends on the task requirement. 
For answer-oriented exploration, the localization stage only localizes a candidate evidence region $r$ and does not require the VLM to answer the question; thus, the evidence content $\mathbf{y}$ is left empty. The answer content is instantiated during verification, where the selected region is examined to determine whether it provides sufficient visual support for answering.
For goal-oriented navigation, the evidence must be grounded to an executable target. \frkabbr therefore instantiates $\mathbf{y}$ as either a verified object instance or an evidential visual point. When the evidence is an object, the instance itself provides semantic identity and spatial localization, and the region field $r$ is omitted. When the evidence is a visual point, \frkabbr retains both $r$ and $\mathbf{y}$: $r$ specifies the coarse evidence region, while $\mathbf{y}$ denotes the detector-independent point to be projected and locally verified.
We instantiate localization and verification functions for different objectives, and describe how their evidence proposals are generated and verified.

\subsubsection*{Answer-Oriented Exploration}

For answer-oriented tasks, \frkabbr performs regional localization over the snapshot memory. 
Since these tasks only need reliable visual support rather than a 3D target pose, \frkabbr localizes evidence at a coarse regional level.
Specifically, each snapshot $S_k\in\mathcal{S}_t$ is divided into a $3\times3$ grid. The VLM screens for regions that may contain question-relevant evidence:
\begin{equation}
    \widehat{\mathcal{E}}_t^{\mathrm{ans}}
    =G_{\mathrm{loc}}^{\mathrm{ans}}(q,\mathcal{S}_t)
    =\{\widehat{e}_k^{\mathrm{ans}}=(S_k,r_k,\varnothing)\}
    ,
    \label{eq:region_screening}
\end{equation}
where $r_k\in\{0,\ldots,8\}$ denotes the selected candidate grid in snapshot $S_k$, and $\varnothing$ indicates that no answer has been verified yet. This coarse regional screening prevents the VLM from answering solely from a whole-image impression. 

Each proposed region is then enlarged to preserve local context and cropped for a stricter verification query:
\begin{equation}
    \mathcal{E}_t^{\mathrm{ans}}
    =G_{\mathrm{ver}}^{\mathrm{ans}}(q,\mathrm{Crop}^{+}(S_k,r_k))
    =\{e_k^{\mathrm{ans}}=(S_k,r_k,a_k)\},
    \label{eq:crop_verify}
\end{equation}
where $\mathrm{Crop}^{+}$ denotes the enlarged crop around region $r_k$, and $a_k$ is the predicted and verified answer. 
Unlike direct whole-snapshot answering, this verification step requires the VLM to judge whether the localized crop itself contains enough visual evidence before answering the question, thereby reducing reliance on coarse scene-level priors.

When multiple regional crops pass verification, \frkabbr performs answer arbitration over the verified evidence by comparing their predicted answers and supporting crops. The agent commits to an answer only when the verified evidence is sufficiently consistent. Otherwise, the current observation is treated as insufficient, and \frkabbr proceeds to frontier-guided exploration. This design enables accurate termination, avoiding early answering from weak global impressions, while also stopping exploration once decisive evidence emerges.

\subsubsection*{Goal-Oriented Exploration}
\label{sec:goal_probing}

For goal-oriented tasks, evidence must be spatially executable because task completion is evaluated by the distance to the target. \frkabbr therefore first instantiates evidence as object-level proposals from the object memory and snapshot memory:
\begin{equation}
    \widehat{\mathcal{E}}_t^{\mathrm{obj}}
    =G^{\mathrm{obj}}_{\mathrm{loc}}(q,\mathcal{S}_t,\mathcal{O}_t)
    =\{\widehat{e}_{k,j}^{\mathrm{obj}}=(S_k,\varnothing,o_j)\},
    \label{eq:object_evidence}
\end{equation}
where $o_j$ is a persistent object instance. 
Specifically, $G^{\mathrm{obj}}_{\mathrm{loc}}$ uses the object crops and their surrounding snapshot context to identify object-snapshot pairs that may match the target.

A selected object is not immediately regarded as decisive evidence. Instead, \frkabbr locks it as a provisional target, navigates toward it, and performs verification using close-range observations. The verifier can confirm the locked object, switch to a better nearby candidate, or reject the proposal. This step prevents a distant snapshot match from being incorrectly used for termination when the object is partially visible or confused with a nearby instance. After verification, the accepted object evidence is 
\begin{equation}
\mathcal{E}_t^{\mathrm{obj}}
=G_{\mathrm{ver}}^{\mathrm{obj}}(q, S_k, o_j)=
\{e_{k,j}^{\mathrm{obj}}=(\tilde{S}_k,\varnothing,\tilde{o}_j)\},
\end{equation}
where $\tilde{o}_j$ and $\tilde{S}_k$ denote the accepted object and the corresponding snapshot. 
If no object is verified locally, the proposal is rejected and cannot trigger task completion.

Object-level evidence, however, depends on the detector and tracking pipeline. The target may be missed, fragmented, or absent from $\mathcal{O}_t$ even when it is visible in the raw image. To recover such missed evidence, \frkabbr introduces an \emph{evidential visual point}, a detector-independent evidence unit predicted directly from recent observations. Unlike persistent object instances, a visual point has no object identity and serves only as a provisional pixel-level cue. 
Each observation is similarly divided into a $3\times 3$ grid. The VLM localizes a task-relevant region and predicts a normalized point within that region:
\begin{equation}
    \widehat{\mathcal{E}}_t^{\mathrm{vp}}
    =G_{\mathrm{loc}}^{\mathrm{vp}}(q,\mathcal{I}_t)
    =\{\widehat{e}_k^{\mathrm{vp}}=(I_k,r_k,\mathbf{u}_k)\},
    \label{eq:visual_point_recall}
\end{equation}
where $r_k\in\{0,\ldots,8\}$ is the selected grid and $\mathbf{u}_k=(u_k,v_k)$ is a normalized image point constrained within that region. Before projection, $\mathbf{u}_k$ is converted to the pixel coordinate
$\bar{\mathbf{u}}_k=(\bar{u}_k,\bar{v}_k)$ according to the image width $W_k$ and height $H_k$:
\begin{equation}
    \bar{\mathbf{u}}_k =
    \left[
    u_k(W_k-1),\,
    v_k(H_k-1)
    \right]^\top .
\end{equation}

The visual point is then back-projected to the 3D camera coordinate system and transformed to the world coordinate system using the depth map $D_k$, camera intrinsics $\mathbf{K}$, and the camera-to-world pose transformation matrix $\mathbf{T}_{k}^{c\rightarrow w}$:
\begin{equation}
    \bar{\mathbf{z}}_k =
    \mathbf{T}_{k}^{c\rightarrow w}
    \begin{bmatrix}
    D_k(\bar{\mathbf{u}}_k)\mathbf{K}^{-1}
    [\bar{u}_k,\bar{v}_k,1]^\top \\
    1
    \end{bmatrix}.
    \label{eq:point_projection}
\end{equation}
where $\bar{\mathbf{z}}_k$ is the homogeneous world coordinate, and $\mathbf{z}_k$ denotes its 3D Euclidean coordinate after dehomogenization.

Since VLM-selected points may drift under viewpoint changes or small-object ambiguity, \frkabbr does not terminate from the projected point alone. After navigating toward $\mathbf{z}_k$, the agent verifies the target again from close-range views and, when possible, obtains the accepted visual point $\tilde{\mathbf{u}}_k$: 
\begin{equation}
    \mathcal{E}_t^{\mathrm{vp}}=G_{\mathrm{ver}}^{\mathrm{vp}}(q,  \mathcal{I}^{\mathrm{near}}_t|_{(I_k, r_k, \mathbf{u}_k)}) = \{e_k^{\mathrm{vp}}=(\tilde{I}^{\mathrm{near}}_k, r_k,\tilde{\mathbf{u}}_k)\},
    \label{eq:visual_point_local_verify}
\end{equation}
where $\mathcal{I}^{\mathrm{near}}_t|_{(I_k, r_k, \mathbf{u}_k)}$ denotes observations after navigating to $\mathbf{z}_k$. 
Otherwise, the point remains an unaccepted proposal.
In this way, objects and evidential visual points provide complementary forms of goal-oriented evidence.

\subsection{Cross-Temporal Frontier Association}
\label{sec:frontier_association}

\hptwognr requires the same frontier to remain identifiable across decision steps. However, frontiers are transient geometric entities whose locations and boundaries may shift, split, or merge as the spatial map evolves. Directly attaching exploration status to instantaneous frontier identities would therefore fragment the feedback accumulated from repeated visits to the same direction. To establish cross-temporal continuity, \frkabbr associates successive frontier instances through persistent spatial-direction keys and maintains their persistent status in a frontier status memory $\mathcal{R}_t$.

\textbf{Spatial-Direction Association:}
Each frontier $F_t^i$ is assigned a key
\begin{equation}
    c_t^i=\left(
    \mathrm{Quant}_{p}(\mathbf{f}_t^i),
    \mathrm{Quant}_{d}(\mathbf{d}_t^i)
    \right),
    \label{eq:key_discr}
\end{equation}
where $\mathbf{f}_t^i$ and $\mathbf{d}_t^i$ denote the frontier centroid position and facing direction, respectively. $\mathrm{Quant}_{p}(\cdot)$ and $\mathrm{Quant}_{d}(\cdot)$ discretize the two attributes such that spatially proximate frontiers with similar orientations can be associated with a common persistent key. This association allows exploration feedback attached to a transient frontier instance to be retrieved when the corresponding direction reappears at subsequent steps.

\textbf{Persistent Frontier Status Memory:}
Based on these persistent keys, \frkabbr maintains the frontier status memory
\begin{equation}
    \mathcal{R}_t=\{c \mapsto (\ell^c,z^c)\},
    \label{eq:frontier_status_memory}
\end{equation}
where $\ell^c$ denotes the place type associated with $c$ estimated by applying a VLM to the frontier-facing observation of the first-appearing frontier associated with $c$, and $z^c$ denotes its stale count. The former preserves semantic context for subsequent frontier evaluation, whereas the latter records how many consecutive selections of the same direction have failed to yield accepted task evidence. Together, they provide a persistent state representation upon which frontier utility can be re-evaluated across interaction rather than independently reconstructed at each decision step.

\textbf{Frontier Status Updating:}
After a frontier $F_t^{i_\star}$ is selected and executed, the newly acquired observations provide feedback for updating the state associated with its persistent key $c_\star=c_t^{i_\star}$. In particular, the stale count is updated as
\begin{equation}
    z^{c_\star} =
    \begin{cases}
    0, & \text{if decisive evidence is observed},\\
    z^{c_\star}+1, & \text{otherwise},
    \end{cases}
    \label{eq:stale_update}
\end{equation}
such that successful evidence acquisition restores the evidential productivity of the corresponding direction, whereas repeated uninformative selections progressively accumulate negative feedback.
For answer-oriented tasks, a frontier is regarded as revealing decisive evidence if the newly acquired snapshots either expose a question-relevant region or pass verification, such as a crop that directly supports an answer. For goal-oriented tasks, an auxiliary VLM jointly considers the new egocentric observations, navigation goal, and detected object categories to assess whether the frontier has revealed useful target-related evidence.

Through persistent association and state updating, exploration outcomes obtained at one decision step are carried forward to later evaluations of the same direction. This cross-temporal continuity provides the state basis for the subsequent \cptwofull and \cpthreefull.

\subsection{\cptwofull}
\label{sec:method_frontier}

When no decisive evidence is found, \frkabbr treats continued exploration as a search for still-missing information. Each frontier $F_t^i$ is therefore evaluated by both the task relevance of its current view $I(F_t^i)$ and the useful space it may further expose. This prospective estimate preserves potentially informative directions, while leaving their accumulated inspection status to the retrospective regulation that follows.

The first perspective is captured by a reactive task-cue score:
\begin{equation}
    r_t^i = G_{\mathrm{r}}(q,I(F_t^i)) \in [0,1],
    \label{eq:reactive_score}
\end{equation}
This score estimates whether visible content near the frontier is already related to the task.

The second is captured by an active hidden-space score:
\begin{equation}
    h_t^i = G_{\mathrm{h}}(I(F_t^i)) \in [0,1],
    \label{eq:hidden_score}
\end{equation}
which estimates whether the frontier is likely to open a new room, corridor, occluded area, or informative viewpoint. Thus, $r_t^i$ exploits visible task cues, whereas $h_t^i$ preserves prospective exploration when direct evidence has not yet appeared.

\frkabbr combines these two scores to obtain the frontier utility:
\begin{equation}
    U_{t,\mathrm{base}}^{i}
    =\alpha r_t^i +(1-r_t^i)^{\gamma}h_t^i,
    \label{eq:base_frontier_score}
\end{equation}
where $\alpha$ controls the contribution of visible task cues, and $\gamma$ modulates the influence of hidden-space potential under weak direct evidence. This formulation gives priority to frontiers with explicit task-related cues when such cues are visible, while allowing hidden-space potential to dominate when all visible evidence remains insufficient.

\subsection{\cpthreefull}
\label{sec:method_exhaustion}

When the space associated with a frontier has been sufficiently inspected yet remains unproductive, \frkabbr uses its inspection status as explicit spatial feedback. It regulates frontier selection through a hard exhaustion-aware rejection rule that suppresses visually exhausted directions and a soft history-aware penalty that down-weights candidates likely to sustain low-progress exploration. Together, these signals reduce redundant exploration.

\subsubsection{Visual Frontier Rejection}

\frkabbr first maintains a visual coverage memory $\mathcal{C}_t$, which accumulates the regions that have been visually inspected by ray-casting egocentric observations at the pose $\mathbf{x}_t$ into the spatial map:
\begin{equation}
    \mathcal{C}_t=\mathcal{C}_{t-1}\cup
    \mathrm{RayCast}(\mathcal{I}_t,\mathbf{x}_t).
    \label{eq:visual_coverage}
\end{equation}
For each frontier $F_t^i$, \frkabbr estimates how much of its associated frontier-facing region (represented by its corresponding spatial mask $m(F_t^i)$) has already been observed:
\begin{equation}
    \kappa_t^i=\frac{|m(F_t^i)\cap\mathcal{C}_t|}{|m(F_t^i)|+\epsilon}.
    \label{eq:frontier_seen_ratio}
\end{equation}
$\kappa_t^i$ measures visual exhaustion rather than geometric visitation. A high value indicates that the region behind or around the frontier has already been inspected from previous viewpoints.

\textbf{Frontier Rejection:} 
For each frontier candidate $F_t^i$, we first use the stale feedback $z_t^{c_i}$ retrieved from $\mathcal{R}_t$ to weaken its reactive task-cue score:
\begin{equation}
    \tilde{r}_t^i = r_t^i\cdot \max(\rho_{\min},1-\delta z_t^{c_i}),
    \label{eq:effective_task_cue}
\end{equation}
where $r_t^i$ is the reactive score, $\tilde{r}_t^i$ is its stale-adjusted value, $\delta$ controls the decay rate, and $\rho_{\min}$ retains a minimum fraction of the original cue to avoid overreacting to sparse feedback. A frontier with an initially strong visible cue can be down-weighted if repeated visits fail to yield accepted evidence.

Rejection is applied when multiple exhaustion signals agree:
\begin{equation}
\begin{split}
    R_t^i = &
    \mathbb{I}(z_t^{c_i}\ge N_z)
    \mathbb{I}(\kappa_t^i\ge\psi_{\mathrm{cov}}) \\
     & \mathbb{I}(\tilde{r}_t^i<\psi_{\mathrm{task}})
    \mathbb{I}(\ell_t^{c_i}\in\mathcal{P}_{\mathrm{sup}}).
    \label{eq:visual_reject}
\end{split}
\end{equation}
$N_z$ is the stale-count threshold, $\psi_{\mathrm{cov}}$ is the visual-coverage threshold, and $\psi_{\mathrm{task}}$ is the minimum effective task-cue threshold. The set $\mathcal{P}_{\mathrm{sup}}$ contains locally suppressible contexts, such as visually exhausted internal rooms or closed regions. This rule is deliberately conservative: a frontier is hard-rejected only when it has been sufficiently observed, repeatedly failed to reveal accepted evidence, retains only a weak effective task cue, and belongs to a context unlikely to open new informative space. Directions that still suggest exits, corridors, open continuations, or unexplored space are therefore preserved, reducing the risk of prematurely blocking exploration.

\subsubsection{History-Aware Low-Progress Penalization}

\frkabbr applies a soft behavioral penalty to discourage frontier choices that continue recent low-progress exploration. This penalty is derived from the frontier-driven decision history:
\begin{equation}
    \mathcal{H}_t=\{\eta_\tau=(\mathbf{p}_\tau^{s},\mathbf{p}_\tau^{e})\}_{\tau<t},
\end{equation}
where $\mathbf{p}_\tau^{s}$ and $\mathbf{p}_\tau^{e}$ denote the start and end positions of a previous frontier decision. For each frontier, \frkabbr computes
\begin{equation}
    P_{t,\mathrm{hist}}^{i}
    =w_{\mathrm{prog}}P_{t,\mathrm{prog}}^{i},
    \label{eq:history_penalty}
\end{equation}
where $P_{t,\mathrm{prog}}^{i}$ penalizes candidates that are likely to keep the agent in a recent low-progress frontier-driven pattern, and $w_{\mathrm{prog}}\ge0$ controls its strength.

For a candidate frontier $F_t^i$, let $\mathbf{q}_i$ denote the next reachable position planned for it, and let $\mathbf{p}_t$ denote the current agent position. To measure whether a candidate remains close to recent frontier outcomes, we define a smooth proximity score:
\begin{equation}
    \phi_R(\mathbf{a},\mathbf{b})
    =\exp\left(-\frac{\|\mathbf{a}-\mathbf{b}\|_2^2}{R^2}\right),
    \label{eq:near_kernel}
\end{equation}
where $\mathbf{a}$ and $\mathbf{b}$ denote two positions, $\|\cdot\|_2$ is the Euclidean distance, and $R$ controls the spatial neighborhood size.

The low-progress signal is estimated from a recent ordered window of frontier decisions, $\mathcal{W}_t=\{\eta_m,\ldots,\eta_n\}\subset\mathcal{H}_t$. Its traveled path length and net displacement are defined as
\begin{equation}
    L(\mathcal{W}_t)=
    \sum_{\eta_j\in\mathcal{W}_t}\|\mathbf{p}_j^{e}-\mathbf{p}_j^{s}\|_2,
    \quad
    \Delta(\mathcal{W}_t)=
    \|\mathbf{p}_n^{e}-\mathbf{p}_m^{s}\|_2 .
    \label{eq:progress_length}
\end{equation}
$L(\mathcal{W}_t)$ measures the navigation budget spent in the recent window, while $\Delta(\mathcal{W}_t)$ measures the net displacement achieved by these decisions. \frkabbr then computes a low-progress score:
\begin{equation}
    \chi_t =
    \left[ 1 - \frac{\Delta(\mathcal{W}_t)/(L(\mathcal{W}_t)+\epsilon)}
    {\beta}
    \right]_0^1,
    \label{eq:global_low_progress}
\end{equation}
where $\beta$ is desired minimum progress ratio, $\epsilon$ is a small constant for numerical stability, and $[\cdot]_0^1$ clips value to $[0,1]$. A larger $\chi_t$ indicates that recent frontier decisions have consumed path length without producing sufficient net displacement.
The candidate-specific penalty is then defined as
\begin{equation}
    P_{t,\mathrm{prog}}^{i}
    =
    \chi_t
    \max\left(
    \max_{\eta_j\in\mathcal{W}_t}\phi_R(\mathbf{q}_i,\mathbf{p}_j^{e}),
    \phi_R(\mathbf{q}_i,\mathbf{p}_t)
    \right).
    \label{eq:low_progress_penalty}
\end{equation}
The penalty is active only when the recent frontier trajectory exhibits poor global progress. It down-weights candidate frontiers that would keep the agent near recent endpoints or its current position, discouraging oscillatory or locally stagnant exploration without permanently blocking the frontier.

Finally, \frkabbr combines hard rejection and soft penalization into the utility:
\begin{equation}
    U_t^i =
    \begin{cases}
    -\infty, & R_t^i=1,\\
    U_{t,\mathrm{base}}^{i}-P_{t,\mathrm{hist}}^{i}, & R_t^i=0.
    \end{cases}
    \label{eq:final_frontier_score}
\end{equation}
The frontier $F_t^{i_\star}$ with the maximum $U_t^i$ is selected for the next-step navigation.

\section{Experiments}
\label{sec:experiments}

We evaluate \frkabbr on main embodied exploration tasks, including active embodied question answering, open-vocabulary single-goal navigation, and long-horizon multi-goal navigation, to examine its effectiveness across diverse exploration settings. The experimental setup is introduced first, followed by comparisons with state-of-the-art methods. We then conduct progressive-addition and leave-one-component-out ablations to assess the contributions of the core mechanisms, and further evaluate generalization across different VLMs.

\subsection{Experimental Setup}

\subsubsection*{Datasets}
Following previous works~\cite{yang20253d,chen2026hgr}, we evaluate \frkabbr on three embodied exploration benchmarks, A-EQA~\cite{majumdar2024openeqa} (the most adopted 184-subset and 41-subset) for embodied QA evaluation, HM3D-OVON~\cite{yokoyama2024hm3d} and GOAT-Bench~\cite{khanna2024goat} validation-unseen full set and 278-subset for single- or multi-goal navigation evaluation. 
For A-EQA, we follow OpenEQA~\cite{majumdar2024openeqa} and report LLM-Match and LLM-Match Success weighted by Path Length (LLM-Match SPL). For HM3D-OVON and GOAT-Bench, we report Success Rate (SR) and SPL. All metrics are reported as percentages.

\subsubsection*{Implementation Details}

\frkabbr uses YOLO-World~\cite{cheng2024yolo} for open-vocabulary object detection, SAM~\cite{kirillov2023segment} for mask extraction, and CLIP~\cite{radford2021learning} features for object association and memory maintenance. GPT-4o is the default VLM choice for all compared methods using the OpenAI API. All experiments are conducted on the NVIDIA 4090 GPUs. Code is \href{https://github.com/Kenneth-Wong/ace}{available}.


Following the default settings~\cite{yang20253d,wang2026scope,lu2026gsmem}, the agent operates in Habitat~\cite{savva2019habitat} with RGB-D observations at $1280\times1280$ resolution, a camera height of $1.5$~m, a downward camera tilt of $30^\circ$, and a horizontal field of view of $120^\circ$. Images sent to VLM prompts are resized to $360\times360$. The agent updates a TSDF map with a voxel size of $0.1$~m and a depth integration range of $1.7$~m. For HM3D-OVON and GOAT-Bench, success is evaluated under the standard $1.0$~m target-distance threshold. 

The \frkabbr-specific hyper-parameters are fixed across experiments. For prospective frontier scoring, we set $\alpha=1.6$ and $\gamma=2$. For \cpthree, we use $N_z=1$, $\psi_{\mathrm{cov}}=0.35$, $\psi_{\mathrm{task}}=0.35$, $\delta=0.35$, and $\rho_{\min}=0.10$. Visual coverage is accumulated by ray-casting observations within $3.5$~m, and the frontier feedback cache uses a $5$-voxel position bin, a $30^\circ$ direction bin, and a 20-step lifetime. For history-aware low-progress penalization, we set $w_{\mathrm{prog}}=0.30$, $\beta=0.35$, $|\mathcal{W}_t|=4$, and $R=8.0$~m, and activate the low-progress signal only after the accumulated frontier-driven path length exceeds $10.0$~m.

\subsection{Quantitative Results on A-EQA}
\label{sec:exp_aeqa_quant_results}

\subsubsection*{Baselines and Compared Methods} 
We compare \frkabbr with representative A-EQA baselines and recent state-of-the-art VLM-based embodied exploration methods. The blind GPT-4o baseline answers questions without visual observations, testing how far language priors alone can support open-ended answering. The question-agnostic Multi-Frame baseline from OpenEQA~\cite{majumdar2024openeqa} explores by generic frontier expansion, stores episodic frames, and prompts the VLM with the collected visual context, representing passive visual accumulation without task-conditioned evidence regulation.

We further compare with methods that actively guide exploration or organize observations into structured memory. Explore-EQA~\cite{ren2024explore} follows a confidence-driven stopping strategy, where the agent answers from the egocentric view once VLM confidence exceeds a preset threshold. ConceptGraphs with frontier exploration~\cite{gu2024conceptgraphs} (CG + Frontier) represents an object-centric memory variant, where object crops from ConceptGraphs are used as memory inputs. Memory-centric zero-shot methods, including 3D-Mem~\cite{yang20253d}, GSMem~\cite{lu2026gsmem}, and BSC-NAV~\cite{ruan2026brain}, maintain structured spatial or persistent 3D memories for VLM reasoning. Experience-enhanced methods, including  ReEXplore~\cite{zhang2025reexplore}, MetaNav~\cite{metanav2026}, and HGR~\cite{chen2026hgr}, incorporate cross-episode retrospective experience, meta-level navigation control, or hypothesis-graph refinement to improve exploration. We also include SCOPE~\cite{wang2026scope}, a recent semantic-cognition and potential-based exploration framework.

\subsubsection*{Results Analysis} 
In \cref{tab:aeqa_main}, \frkabbr achieves the best performance. It improves the strongest prior LLM-Match result from 59.1\% to 60.4\%, and more importantly raises LLM-Match SPL to 50.2\%, surpassing the best previous value of 45.5\% from MetaNav by 4.7\%. It indicates that \frkabbr improves not only answer quality, but also the  timing of termination.

The comparison further shows that efficient exploration cannot be achieved by stronger memory or history regulation alone. ReEXplore improves answer matching through retrospective episode experience, but its SPL remains 37.3\%, suggesting that experience-enhanced reasoning does not necessarily yield efficient evidence acquisition in the current episode. MetaNav obtains the strongest prior SPL with history-aware navigation regulation, but still trails \frkabbr. By verifying whether localized visual evidence is sufficient before termination, \frkabbr avoids both unsupported early answers and unnecessary continued search, while its frontier evaluation and feedback regulation reduce unproductive exploration.

\begin{table}[ht]
\renewcommand{\arraystretch}{1.1}
\centering
\caption{Performance comparison on A-EQA 184-subset. The best and second-best results are highlighted in bold and underlined, respectively.}
\label{tab:aeqa_main}
\setlength{\tabcolsep}{9pt}
\begin{tabular}{@{}lcc@{}}
\toprule
Method & LLM-Match (\%) $\uparrow$ & LLM-Match SPL (\%) $\uparrow$ \\

\multicolumn{3}{@{}l}{\textit{\textbf{Baselines}}} \\
GPT-4o~\cite{yang20253d} & 35.9 & -- \\
Multi-Frame*~\cite{majumdar2024openeqa} & 41.8 & 7.5 \\
\midrule
\multicolumn{3}{@{}l}{\textit{\textbf{VLM-based exploration}}} \\
Explore-EQA~\cite{ren2024explore} & 46.9 & 23.4 \\
CG + Frontier~\cite{yang20253d,gu2024conceptgraphs} & 47.2 & 33.3 \\
3D-Mem~\cite{yang20253d} & 52.6 & 42.0 \\
BSC-NAV~\cite{ruan2026brain} & 54.6 & -- \\
ReEXplore~\cite{zhang2025reexplore} & 58.3 & 37.3 \\
MetaNav~\cite{metanav2026} & 58.3 & \underline{45.5} \\
GSMem~\cite{lu2026gsmem} & 55.4 & 43.8 \\
HGR~\cite{chen2026hgr} & 55.9 & 45.0 \\
SCOPE~\cite{wang2026scope} & \underline{59.1} & 41.0 \\
\textbf{\frkabbr (Ours)} &  \textbf{60.4} & \textbf{50.2} \\
\bottomrule
\end{tabular}
\end{table}

\subsection{Quantitative Results on HM3D-OVON}

\subsubsection*{Compared Methods} 
On HM3D-OVON, we compare \frkabbr with both supervised and training-free open-vocabulary navigation methods. The supervised group includes behavior cloning (BC)~\cite{pomerleau1988alvinn}, DAgger-style training~\cite{ross2011reduction}, object-detector-enhanced navigation~\cite{yokoyama2024hm3d}, and recent foundation-model-based policies~\cite{zhang2024uni,zemskova2026ovsegdt,zhu2025mtu,wang2026dynam3d,zhang2025embodied}. These methods serve as strong references but generally depend on task-specific adaptation. The training-free group includes VLFM~\cite{yokoyama2023vlfm}, TANGO~\cite{ziliotto2025tango}, BSC-NAV~\cite{ruan2026brain}, and MetaNav~\cite{metanav2026}. They are the most direct comparisons to \frkabbr, as they use foundation models without training a policy on the target benchmark.

\subsubsection*{Results Analysis} 
In \cref{tab:hm3dovon_main}, \frkabbr achieves 54.4\% SR and 32.8\% SPL. Compared with the strongest training-free SR baseline, MetaNav, \frkabbr improves SR by 8.3\%. Compared with BSC-NAV, which obtains the strongest prior SPL, \frkabbr slightly improves SPL while increasing SR by 14.2\%.

The improvement pattern differs from A-EQA. On A-EQA, the gain is more pronounced on LLM-Match SPL than on LLM-Match, because many recent VLM-based methods already produce competitive answers, and \frkabbr mainly improves the efficiency. On HM3D-OVON, however, the gain is more pronounced on SR than on SPL. This is because goal navigation requires the agent to physically localize a target instance, and success is often limited by whether the correct object can be discovered. Once a target is found, the remaining path efficiency is strongly constrained by scene topology, target distance, and object accessibility. Therefore, \frkabbr primarily improves reliable target grounding through object evidence verification and prospective frontier evaluation, while preserving rather than dramatically increasing SPL.

\begin{table}[ht]
\renewcommand{\arraystretch}{1.1}
\centering
\caption{Performance comparison on HM3D-OVON.}
\label{tab:hm3dovon_main}
\setlength{\tabcolsep}{31pt}
\begin{tabular}{@{}lcc@{}}
\toprule
Method & SR (\%) $\uparrow$ & SPL (\%) $\uparrow$ \\
\midrule
\multicolumn{3}{@{}l}{\textit{\textbf{Supervised methods}}} \\
BC~\cite{pomerleau1988alvinn}  & 5.4 & 1.9 \\
DAgRL~\cite{ross2011reduction} & 18.3 & 7.9 \\
DAgRL+OD~\cite{yokoyama2024hm3d} & 37.1 &  19.8 \\
Uni-NaVid~\cite{zhang2024uni} & 39.5 & 19.8 \\
OVSegDT~\cite{zemskova2026ovsegdt} & 40.1 & 20.9 \\
MTU3D~\cite{zhu2025mtu} & 40.8 & 12.1 \\
Dynam3D~\cite{wang2026dynam3d} & 42.7 & 22.4 \\
NavFoM~\cite{zhang2025embodied} & 45.2 & 31.9 \\
\multicolumn{3}{@{}l}{\textit{\textbf{Training-free methods}}} \\
VLFM~\cite{yokoyama2023vlfm} & 35.2 & 19.6 \\
TANGO~\cite{ziliotto2025tango} & 35.5 & 19.5 \\  
BSC-NAV~\cite{ruan2026brain} & 40.2 & \underline{32.5} \\
MetaNav~\cite{metanav2026} & \underline{46.1} & 29.8 \\
\textbf{\frkabbr (Ours)} &  \textbf{54.4} & \textbf{32.8} \\
\bottomrule
\end{tabular}
\end{table}

\subsection{Quantitative Results on GOAT-Bench}

\subsubsection*{Compared Methods}
For GOAT-Bench, we compare \frkabbr with supervised policies and training-free navigation systems. The supervised group includes SenseAct-NN Skill Chain and SenseAct-NN Monolitic~\cite{khanna2024goat}, which respectively use chained navigation skills and a single goal-conditioned policy for long-horizon execution. We also compare with recently learned 3D navigation methods, including MTU3D~\cite{zhu2025mtu} and AstraNav-Memory~\cite{hu2026astranav}, the latter emphasizing compressed long-term memory for extended navigation.
The training-free group includes modular systems and recent VLM-based exploration methods. Modular GOAT and Modular Clip on Wheels~\cite{khanna2024goat} combine perception, mapping, and planning components without end-to-end policy learning. We further compare with TANGO~\cite{ziliotto2025tango}, ConceptGraph-based methods~\cite{gu2024conceptgraphs} (CG + Frontier), 3D-Mem~\cite{yang20253d}, MetaNav~\cite{metanav2026}, HGR~\cite{chen2026hgr}, SCOPE~\cite{wang2026scope}, and GSMem~\cite{lu2026gsmem}. These methods provide strong references for zero-shot long-horizon exploration with external perception, structured memory, VLM reasoning, and history- or hypothesis-guided navigation.

\subsubsection*{Results Analysis} 
In \cref{tab:goat_full_subset}, under the task setting of long-horizon multimodal goals,  \frkabbr still achieves the best SR on both the full set and the 278-subset, reaching 69.9\% and 76.3\% SR respectively, substantially better than previous SOTA methods. It also achieves competitive path efficiency among training-free methods, with an SPL second only to HGR on the full set. 
The improvement is more pronounced in SR than in SPL, which is similar to that on HM3D-OVON.

\begin{table}[t]
\renewcommand{\arraystretch}{1.1}
\centering
\caption{Performance comparison on the GOAT-Bench. }
\label{tab:goat_full_subset}
\setlength{\tabcolsep}{9pt}  
\begin{tabular}{@{}lcccc@{}}
\toprule
Method &
\multicolumn{2}{c}{Full set} &
\multicolumn{2}{c}{278-subset} \\
\cmidrule(lr){2-3}\cmidrule(lr){4-5}
& SR $\uparrow$ & SPL $\uparrow$ & SR $\uparrow$ & SPL $\uparrow$ \\
\midrule
\multicolumn{3}{@{}l}{\textit{\textbf{Supervised methods}}} \\
SenseAct-NN Skill Chain~\cite{khanna2024goat} & 29.5 & 11.3 & -- & -- \\
SenseAct-NN Monolitic~\cite{khanna2024goat} & 12.3 & 6.8 & -- & -- \\
MTU3D~\cite{zhu2025mtu} & 47.2 & 27.7 & -- & -- \\
AstraNav-Memory~\cite{hu2026astranav} & 62.7 & \textbf{56.9} & -- & -- \\
\midrule
\multicolumn{3}{@{}l}{\textit{\textbf{Training-free methods}}} \\
Modular GOAT~\cite{khanna2024goat} & 24.9 & 17.2 & -- & -- \\
Modular Clip on Wheels~\cite{khanna2024goat} & 12.3 & 10.4 & -- & -- \\
TANGO~\cite{ziliotto2025tango} & 32.1 & 16.5 & -- & --\\  %
Explore-EQA~\cite{ren2024explore} & -- & -- & 55.0 & 37.9 \\
CG + Frontier~\cite{yang20253d,gu2024conceptgraphs} & -- & -- & 61.5 & 41.3 \\
ConceptGraph~\cite{gu2024conceptgraphs,chen2026hgr} & 61.2 & 44.3 & 67.8 & 48.1 \\
3D-Mem w/o memory~\cite{yang20253d} & -- & -- & 58.6 & 38.5 \\ 
3D-Mem~\cite{yang20253d} & 62.9 & 44.7 & 69.1 & 48.9 \\
MetaNav~\cite{metanav2026} & -- & -- & 71.4 & 51.8 \\
HGR~\cite{chen2026hgr} & 64.1 & \textbf{50.1} & 72.4 & \textbf{56.2} \\
SCOPE~\cite{wang2026scope} & 66.8 & 46.5 & \underline{73.7} & \underline{53.5} \\
GSMem~\cite{lu2026gsmem} & \underline{67.2} & 46.9 & -- & -- \\
\textbf{\frkabbr (Ours)} & \textbf{69.9} & \underline{47.7} & \textbf{76.3} & 50.5 \\
\bottomrule
\end{tabular}
\end{table}

\subsection{Ablation Study on Main Components}

We conduct component ablations to examine how the main mechanisms of \frkabbr contribute to different types of embodied exploration. Separate ablations are reported for answer-oriented and goal-oriented tasks because \frkabbr instantiates evidence probing differently under the two task requirements. In A-EQA, the agent only needs sufficient visual support for answering, so \frkabbr localizes regional evidence from snapshot memory and verifies enlarged crops before committing to an answer. In GOAT-Bench and HM3D-OVON, the agent must ground evidence to an executable spatial target. Object candidates provide a natural evidence unit, while evidential visual points further recover detector-independent cues when object evidence is missing or unstable.

\subsubsection*{Baseline Design}
This task-dependent termination design also determines how the ablation baselines should be constructed. 
\cponefull cannot be simply removed, because the agent still needs to decide whether current observations are sufficient for termination or whether exploration should continue. We introduce two simplified baselines with weaker decision strategies. Baseline-1 uses a holistic VLM decision: candidate evidence and frontiers are placed in the same decision context, and the VLM directly selects an executable option. Baseline-2 separates termination from continued exploration using a simplified two-stage decision process. It asks the VLM whether current observations provide task-sufficient support. If not, frontier candidates are selected only by the reactive task-cue score.

\subsubsection*{Ablation Design}
We report two types of ablations. Progressive-addition variants show how performance changes as \frkabbr gradually incorporates \cponefull (\cponeabbr), \cptwofull (\cptwoabbr), and \cpthreefull (\cpthreeabbr). Leave-one-component-out variants measure the performance drop caused by removing each mechanism from the full framework. For \cptwoabbr and \cpthreeabbr, removal is implemented by simplifying the frontier utility, such as disabling hidden-space scoring or removing the corresponding rejection and penalty item.  
We additionally remove the \cponever from \cponeabbr while retaining \cponeloc, to isolate the effect of the localization-to-verification process.

\begin{table*}[!ht]
\renewcommand{\arraystretch}{1.1}
\centering
\caption{
Ablation study of \frkabbr main components on the A-EQA 41-subset. \cponefull (\cponeabbr) refers to region-based \cponeloc and \cponever. \cptwofull (\cptwoabbr) refers to the Active Hidden-Space (AHS) score. \cpthreefull (\cpthreeabbr) includes Visual Frontier Rejection (VFR) and History-aware Low-progress Penalization (HLP). For incremental ablations and baseline comparisons, $\Delta$ is computed relative to the adjacent previous row; for single-component removal ablations, $\Delta$ is computed relative to the full \frkabbr framework.
}
\label{tab:ablation_aeqa_components}
\setlength{\tabcolsep}{11pt}
\resizebox{0.85\textwidth}{!}{
\begin{tabular}{@{}lcccccccc@{}}
\toprule
Method &
\shortstack{\cponeabbr} &
\shortstack{\cptwoabbr (AHS)} &
\multicolumn{2}{c}{\cpthreeabbr} &
\shortstack{LLM-Match $\uparrow$} &
$\Delta$ &
\shortstack{LLM-Match SPL $\uparrow$} &
$\Delta$ \\
\cmidrule(lr){4-5} \cmidrule(lr){6-7}  \cmidrule(lr){8-9}
&  &  & VFR & HLP & & & & \\
\midrule
Baseline-1 (holistic) & -- & -- & $\times$ & $\times$ & 51.4 & -- & 35.1 & --  \\ 
Baseline-2 (separated) & $\times$ & $\times$ & $\times$ & $\times$ & 51.4 & \textcolor{red}{+0.0} & 37.2 & \textcolor{red}{+2.1} \\ 
\midrule
\cponeabbr w/o verify & & $\times$ & $\times$ & $\times$ & 53.1 & \textcolor{red}{+1.7} & 39.9 & \textcolor{red}{+2.7} \\
\cponeabbr & $\checkmark$ & $\times$ & $\times$ & $\times$ & 56.5 & \textcolor{red}{+3.4} & 41.8 & \textcolor{red}{+1.9} \\
\cponeabbr+AHS & $\checkmark$ & $\checkmark$ & $\times$ & $\times$ & 57.6 & \textcolor{red}{+1.1} & 41.1 & \textcolor{forestgreen}{-0.7} \\
\cponeabbr+AHS+VFR & $\checkmark$ & $\checkmark$ & $\checkmark$ & $\times$ & 59.3 & \textcolor{red}{+1.7} & 44.8 & \textcolor{red}{+3.7} \\
\cponeabbr+AHS+VFR+HLP (full \frkabbr) & $\checkmark$ & $\checkmark$ & $\checkmark$ & $\checkmark$ & 61.6 & \textcolor{red}{+2.3} & 47.2 & \textcolor{red}{+2.4} \\\midrule
w/o \cponeabbr & $\times$ & $\checkmark$ & $\checkmark$ & $\checkmark$ & 55.2 & \textcolor{forestgreen}{-6.4} & 42.0 & \textcolor{forestgreen}{-5.2} \\
w/o AHS & $\checkmark$ & $\times$ & $\checkmark$ & $\checkmark$ & 59.9 & \textcolor{forestgreen}{-1.7} & 46.1 & \textcolor{forestgreen}{-1.1} \\
w/o VFR & $\checkmark$ & $\checkmark$ & $\times$ & $\checkmark$ & 60.6 & \textcolor{forestgreen}{-1.0} & 43.3 & \textcolor{forestgreen}{-3.9} \\
w/o HLP & $\checkmark$ & $\checkmark$ & $\checkmark$ & $\times$ & 59.3 & \textcolor{forestgreen}{-2.3} & 44.8 & \textcolor{forestgreen}{-2.4} \\
\bottomrule
\end{tabular}}
\end{table*}

\subsubsection*{Ablation on A-EQA}
In \cref{tab:ablation_aeqa_components}, the two baselines reveal the limitation of holistic VLM decision-making. Baseline-2 improves LLM-Match SPL from 35.1\% to 37.2\% over Baseline-1, but its LLM-Match remains unchanged at 51.4\%. This indicates that a simple ``judge-then-explore'' protocol can reduce some inefficient decisions, but it does not provide stronger evidence. 
Introducing \cponeabbr without verification improves LLM-Match and SPL to 53.1\% and 39.9\%. Verification raises them to 56.5\% and 41.8\%, confirming that localization-to-verification is critical for converting plausible cues into decisive evidence. Overall, \cponeabbr brings the largest improvement, confirming that A-EQA benefits most from evidence localization and verification rather than coarse whole-view judgment or direct action selection.

The remaining components regulate where further evidence should be sought and which directions should be deprioritized. Adding AHS improves LLM-Match from 56.5\% to 57.6\%, showing that prospective frontier evaluation can expose additional evidence. Its slight SPL drop from 41.8\% to 41.1\% suggests that seeking hidden space may sometimes induce longer exploration when used without rejection control. After VFR is introduced, LLM-Match SPL increases notably to 44.8\%, indicating that suppressing visually exhausted frontier directions helps convert prospective search into more efficient evidence acquisition. HLP improves the full \frkabbr model to 61.6\% LLM-Match and 47.2\% LLM-Match SPL, suggesting that penalizing locally stagnant exploration benefits both answer reliability and task efficiency.

The leave-one-component-out results lead to the same conclusion. Removing \cponeabbr causes the largest degradation, with drops of 6.4\% in LLM-Match and 5.2\% in LLM-Match SPL, confirming that discriminative evidence acceptance is the central mechanism for answer-oriented exploration. Removing AHS reduces both metrics. Removing VFR mainly damages efficiency, causing a 3.9\% SPL drop, while removing HLP also leads to consistent decreases in both answer quality and SPL. Overall, A-EQA relies primarily on granular evidence probing to determine reliable answer commitment.

\begin{table*}[!ht]
\renewcommand{\arraystretch}{1.1}
\centering
\caption{
Ablation study of \frkabbr main components on the GOAT-Bench 278-subset. \cponefull (\cponeabbr) refers to Evidential Visual Point Localization and Verification (EVPLV). 
}
\label{tab:ablation_goat_components}
\setlength{\tabcolsep}{13.5pt}
\resizebox{0.85\textwidth}{!}{
\begin{tabular}{@{}lcccccccc@{}}
\toprule
Method &
\shortstack{\cponeabbr (EVPLV)} &
\shortstack{\cptwoabbr (AHS)} &
\multicolumn{2}{c}{\cpthreeabbr} &
SR $\uparrow$ &
$\Delta$ &
SPL $\uparrow$ &
$\Delta$ \\
\cmidrule(lr){4-5}  \cmidrule(lr){6-7}  \cmidrule(lr){8-9}
& & & VFR & HLP & & & & \\
\midrule
Baseline-1 (holistic) & -- & -- & $\times$ & $\times$ & 68.8 & -- & 41.7 & -- \\
Baseline-2 (separated) & $\times$ & $\times$ & $\times$ & $\times$ & 69.3 & \textcolor{red}{+0.5} & 43.2  & \textcolor{red}{+1.5} \\
\midrule
EVPLV w/o verify & & $\times$ & $\times$ & $\times$ & 70.5 & \textcolor{red}{+1.2} & 44.1 & \textcolor{red}{+0.9} \\
EVPLV & $\checkmark$ & $\times$ & $\times$ & $\times$ & 72.9 & \textcolor{red}{+2.4} & 45.4 & \textcolor{red}{+1.3} \\
EVPLV+AHS & $\checkmark$ & $\checkmark$ & $\times$ & $\times$ & 75.7 & \textcolor{red}{+2.8} & 47.6 & \textcolor{red}{+2.2} \\
EVPLV+AHS+VFR & $\checkmark$ & $\checkmark$ & $\checkmark$ & $\times$ & 76.0 & \textcolor{red}{+0.3} & 49.9 & \textcolor{red}{+2.3} \\
EVPLV+AHS+VFR+HLP (full \frkabbr) & $\checkmark$ & $\checkmark$ & $\checkmark$ & $\checkmark$ & 76.3 & \textcolor{red}{+0.3} & 50.5 & \textcolor{red}{+0.6} \\
\midrule
w/o EVPLV & $\times$ & $\checkmark$ & $\checkmark$ & $\checkmark$ & 73.0 & \textcolor{forestgreen}{-3.3} & 47.1 & \textcolor{forestgreen}{-3.4} \\
 w/o AHS & $\checkmark$ & $\times$ & $\checkmark$ & $\checkmark$ & 73.6 & \textcolor{forestgreen}{-2.7} & 47.9 & \textcolor{forestgreen}{-2.6} \\
 w/o VFR & $\checkmark$ & $\checkmark$ & $\times$ & $\checkmark$ & 76.1 & \textcolor{forestgreen}{-0.2} & 48.4 & \textcolor{forestgreen}{-2.1} \\
 w/o HLP & $\checkmark$ & $\checkmark$ & $\checkmark$ & $\times$ & 76.0 & \textcolor{forestgreen}{-0.3} & 49.9 & \textcolor{forestgreen}{-0.6} \\
\bottomrule
\end{tabular}}
\end{table*}

\subsubsection*{Ablation on Goat-Bench}
In \cref{tab:ablation_goat_components}, the two simplified baselines show that separating termination from exploration is still beneficial in long-horizon navigation. 
EVPLV without verification already improves SR and SPL to 70.5\% and 44.1\%, while adding verification further raises them to 72.9\% and 45.4\%. The larger verification gain highlights its importance for visual-point evidence, whose projected locations are more susceptible to viewpoint variation, localization drift, and small-object ambiguity. Overall, EVPLV helps effectively recover cues that are missed or unstable in object memory.

Adding AHS further improves SR and SPL to 75.7\% and 47.6\%, indicating that hidden-space-aware frontier seeking is particularly important for long-horizon target search. Since GOAT-Bench subtasks often require finding targets beyond the current view, object memory, or explored area, prospective frontier evaluation directly affects whether missing target evidence can be discovered. The leave-one-out results support this observation: removing EVPLV reduces SR and SPL by 3.3\% and 3.4\%, while removing AHS reduces them by 2.7\% and 2.6\%, respectively. Thus, goal-oriented exploration benefits strongly from both detector-independent evidence recovery and active search for unseen target evidence.

The \cpthreeabbr components play a more efficiency-oriented role. Adding VFR produces only a small SR gain, but improves SPL from 47.6\% to 49.9\%. Similarly, removing VFR barely changes SR but causes a 2.1\% SPL drop, showing that VFR mainly prevents excessive visits to exhausted regions. HLP provides a smaller but consistent improvement, raising the full model to 76.3\% SR and 50.5\% SPL, and its removal also slightly decreases both metrics. 

These trends reveal a task-dependent division of labor among \frkabbr components. A-EQA is more sensitive to discriminative evidence acceptance and timely termination. GOAT-Bench is more sensitive to target grounding and active discovery of missing evidence. 
In both settings, frontier regulation improves efficiency by suppressing unproductive exploration, with its benefits reflected more strongly in SPL than in SR.

\subsection{Generalization across VLMs}

We replace GPT-4o with multiple Qwen-family VLMs~\cite{team2026qwen3}. We test Qwen3-VL-32B, Qwen3.5-27B, Qwen3.5-35B, and Qwen3.5-397B on the A-EQA 184-subset, and evaluate Qwen3.5-397B on the GOAT-Bench 278-subset. 
The results in \cref{tab:vlm_replacement} show that \frkabbr maintains competitive performance. Qwen3.5-397B approaches GPT-4o on both A-EQA and GOAT-Bench, and even achieves a higher SPL. Considering its much lower API cost and potential for local deployment, \frkabbr can cooperate with more cost-effective and deployable VLMs while retaining competitive performance.

\begin{table}[ht]
\renewcommand{\arraystretch}{1.15}
\footnotesize
\centering
\caption{Replacement of VLMs. Cost denotes input/output API price per million tokens.}
\label{tab:vlm_replacement}
\setlength{\tabcolsep}{3.0pt}
\begin{tabular}{@{}llccc@{}}
\toprule
Dataset & VLM & Cost & LLM-Match/SR $\uparrow$ &  SPL $\uparrow$ \\
\midrule
\multirow{5}{*}{A-EQA}
& Qwen3-VL-32B & \$0.16/\$0.65 & 50.5 & 37.7 \\
& Qwen3.5-27B & \$0.08/\$0.67 & 54.6 & 45.8 \\
& Qwen3.5-35B & \$0.06/\$0.44 & 53.6 & 49.7 \\
& Qwen3.5-397B & \$0.17/\$1.00 & 56.9 & 53.4 \\
& GPT-4o & \$5.00/\$15.00 & 60.4 & 50.2 \\
\midrule
\multirow{2}{*}{GOAT-Bench}
& Qwen3.5-397B & \$0.17/\$1.00 & 72.4 & 51.1 \\
& GPT-4o & \$5.00/\$15.00 & 76.3 & 50.5 \\
\bottomrule
\end{tabular}
\end{table}

\section{Behavior Analyses}

We conduct behavior analyses to verify whether \frkabbr induces the intended spatially explicit behaviors in termination and frontier evaluation. We first examine observation-side localization and verification, then analyze frontier-side behavior in terms of weak-cue evidence seeking, exhausted-region reselection, and local trajectory progress.

\subsection{Analysis of \cponefull}

Results in \cref{tab:ablation_aeqa_components,tab:ablation_goat_components} have shown that \cponeabbr is a major contributor to improvements in task success. We examine it through more detailed metrics to better understand its effect.

\subsubsection{Answer-Oriented Probing}
\begin{figure}[ht]
\setlength{\belowcaptionskip}{-0.2cm}
    \centering
    \includegraphics[width=\linewidth]{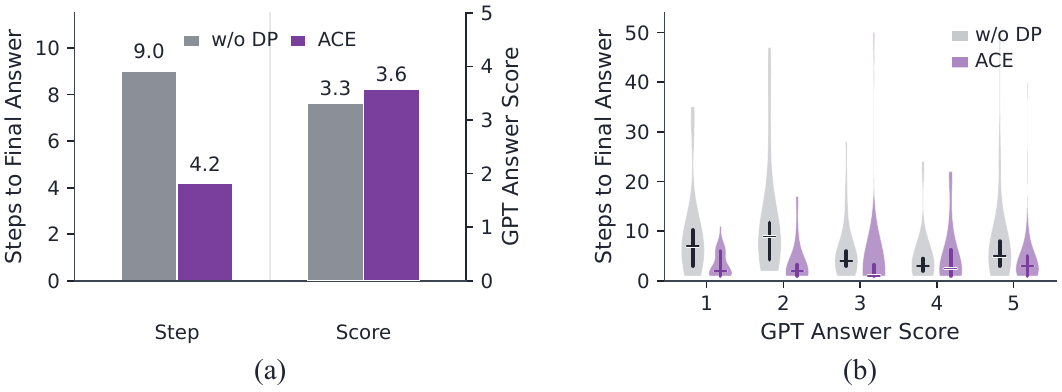}
    \caption{Efficiency--quality statistics of regional evidence probing on A-EQA. 
    (a) Average steps to final answer commitment and average GPT-based answer score for ``w/o \cponeabbr'' and \frkabbr.
    (b) Score-conditioned distributions of commitment steps. Each violin represents the step distribution within an answer-score group. The horizontal bar indicates the median step count, and the vertical bar indicates the interquartile range.
    }
    \label{fig:aeqa_step_scores_distribution_union}
\end{figure}

We evaluate region-based \cponeabbr in A-EQA. Specifically, we compare the ``w/o \cponeabbr'' setting in \cref{tab:ablation_aeqa_components} with the full \frkabbr in terms of the average number of steps before producing the final answer and the final GPT-based answer score, as summarized in \cref{fig:aeqa_step_scores_distribution_union}(a). \cponeabbr reduces the average step count from 9.0 to 4.2 while improving the average answer score from 3.32 to 3.56. This result indicates that \frkabbr does not improve answer quality by exploring longer. Instead, it commits earlier after suggestive evidence has been identified and verified.
We further compare the step distributions under different answer-score groups, as shown in \cref{fig:aeqa_step_scores_distribution_union}(b). Across most score levels, \frkabbr produces more compact distributions with lower step counts, whereas ``w/o \cponeabbr'' exhibits longer tails and more delayed answering. This trend suggests that region-based probing helps avoid unnecessary exploration once decisive evidence has been observed. Overall, the comparison shows that \cponeabbr achieves higher answer quality while requiring fewer steps for commitment.

\begin{figure}[ht]
    \centering
    \includegraphics[width=1.0\linewidth]{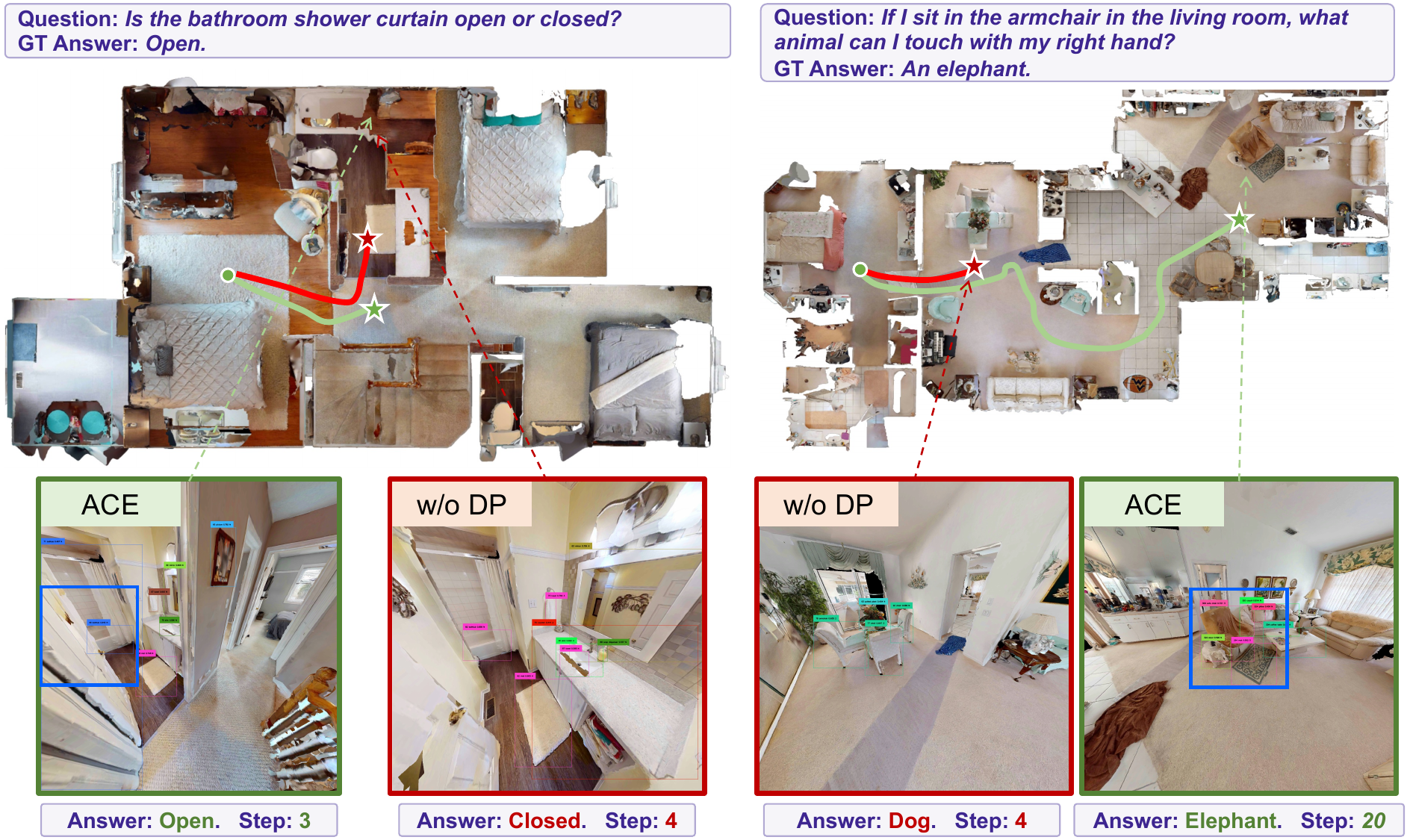}
    \caption{Qualitative comparison of answer commitment. 
Each example shows the explored trajectory (\textcolor{lightgreen}{\frkabbr} vs. \textcolor{red}{w/o DP}), the snapshot used for final answering, the corresponding predicted answer, and step count. The blue box denotes the selected region from the $3\times3$ regional partition, which is identified by the VLM as the evidence-bearing region for answering. }
    \label{fig:exp_region_prob}
\end{figure}

We provide qualitative examples in \cref{fig:exp_region_prob} to illustrate how region-based probing regulates \emph{when} \frkabbr should commit in A-EQA. The key point is not simply that \frkabbr tends to answer earlier, but that it answers \emph{when the available evidence is decisive}. In the left example, both \frkabbr and ``w/o \cponeabbr'' reach highly similar snapshots near the bathroom. However, ``w/o \cponeabbr'' answers \emph{Closed} from the whole-view impression, whereas \frkabbr performs finer regional analysis and correctly focuses on the localized middle-left region (in the blue box), indicating that the shower curtain is \emph{Open}, even though the shower curtain is not detected. This case shows that region-based probing allows \frkabbr to commit early when decisive evidence is visible, while improving answer reliability by grounding the decision in a localized evidence-bearing region.

The right example highlights the complementary case. Although ``w/o \cponeabbr'' terminates after only 4 steps and answers \emph{Dog}, \frkabbr does not accept the current view as task-sufficient, even after briefly exploring around that area. Instead, it continues navigation until reaching the living room specified in the question, where it finally identifies the correct answer \emph{Elephant} at step 20. This case shows that \cponeabbr commits quickly when the answer is already supported, but is also willing to continue exploration when the current evidence is ambiguous.

Taken together, these examples clarify the role of region-based probing in answer-oriented exploration. It enables \frkabbr to answer immediately when decisive evidence emerges, thereby avoiding unnecessary exploration. In this sense, \frkabbr improves both success rate and efficiency by regulating the timing of termination according to evidential sufficiency.

\subsubsection{Goal-Oriented Probing}
As the results in \cref{tab:ablation_goat_components} have shown the significant effect of the evidential visual points (EVPs) on SR and SPL, 
we further focus on how point-level evidence complements object-level evidence.

\begin{figure}[!ht]
    \centering
    \includegraphics[width=\linewidth]{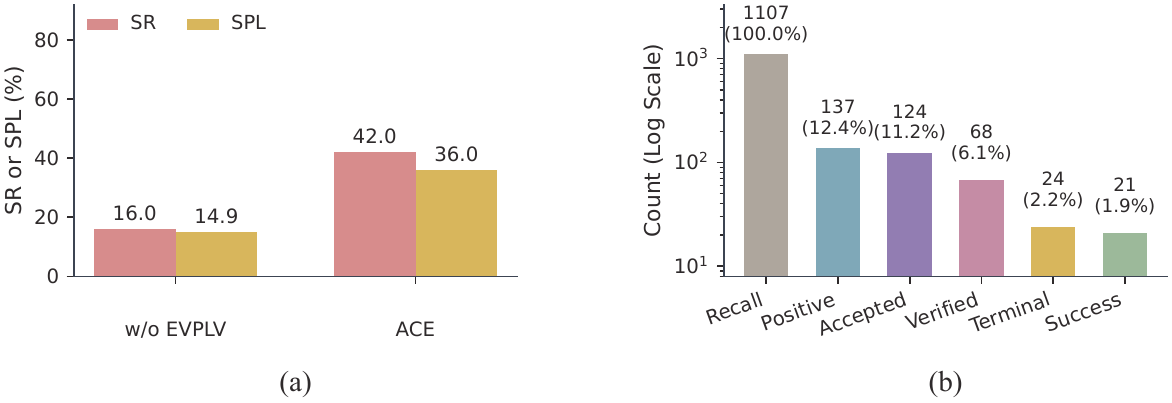}
    \caption{
    Analysis of EVPs.
    (a) The full \frkabbr and w/o EVPLV are compared when no verified object evidence is available. 
    (b) The EVP candidates are progressively filtered.
}
\label{fig:vp_no_object_recovery_and_evidence_funnel}
\end{figure}

We first evaluate the ability of EVP to recover from missing object-level evidence. Under the ``w/o EVPLV'' setting in \cref{tab:ablation_goat_components}, we identify 50 subtasks in which the agent passes near the ground-truth target but fails to establish verified object evidence because the target is either missed by the detector or lost during tracking. Without EVPLV, only 8 subtasks succeed, typically because the agent happens to stop within the success radius. We therefore rerun full \frkabbr on the same 50 subtasks to assess whether EVP can recover the missing target evidence. \frkabbr ultimately obtains a terminal EVP in 24 subtasks and 21 of them succeed. As shown in \cref{fig:vp_no_object_recovery_and_evidence_funnel}(a), these subtasks achieve 42\% SR and 36.0\% SPL, substantially outperforming the corresponding results without EVPLV. This comparison suggests that EVPs provide a useful detector-independent recovery channel when the target cue is visible but cannot be reliably represented as an object target.

We examine the EVP filtering process in \cref{fig:vp_no_object_recovery_and_evidence_funnel}(b). Among 1,107 EVP recall attempts, 137 produce positive proposals, 124 are accepted as navigation cues, 68 pass close-range verification, 24 become terminal EVPs, and 21 finally satisfy the benchmark success criterion. The large reduction from recall attempts to terminal EVPs indicates that the EVPs are filtered conservatively rather than directly used for termination. The high success ratio among terminal EVPs, 21/24, is consistent with the recovery result above and shows that terminal EVP evidence is relatively reliable after verification.

\subsubsection*{Box-Based Evidence}
As an alternative form of evidence localization, we evaluate a bbox-based variant, denoted as \emph{\frkabbr w. bbox}. In this variant, the VLM is asked to localize the candidate target with a bounding box rather than an EVP, while the remaining procedures are kept the same as \frkabbr. This comparison examines whether point-level evidence is merely a different output format, or whether it provides a more suitable localization interface for VLM-based exploration.

As shown in \cref{tab:evp_bbox_comparison}, the bbox-based variant remains competitive, indicating that evidence probing is not tied to a specific localization format. However, its lower SR and SPL suggest that requiring a precise box is not the best interface.

In open-ended egocentric observations, some targets are small, partially visible, irregular in shape, or lack clear object boundaries. Others are better specified by a reference image or contextual description than by a canonical object appearance. Under such conditions, a bounding box can be unstable or overly restrictive. EVP instead provides a lightweight point-level cue that only needs to indicate a plausible target-related location for navigation. Since \frkabbr still performs close-range verification before commitment, the initial point does not need to be a complete localization result. These results suggest that EVP offers a more flexible and spatially actionable evidence carrier for goal-oriented exploration.

\begin{table}[ht]
\renewcommand{\arraystretch}{1.2}
    \centering
    \caption{Comparison between bbox-based localization and EVP localization on the GOAT-Bench 278-subset.}
    \label{tab:evp_bbox_comparison}
    \begin{tabular}{lcc}
        \toprule
        Method & SR $\uparrow$ & SPL $\uparrow$ \\
        \midrule
        \frkabbr w. bbox & 74.2 & 47.7 \\
        \frkabbr & 76.3 & 50.5 \\
        \bottomrule
    \end{tabular}
\end{table}

\subsubsection*{Cases Analysis}
\begin{figure}[ht]
\setlength{\belowcaptionskip}{-0.3cm}
    \centering
    \includegraphics[width=1.0\linewidth]{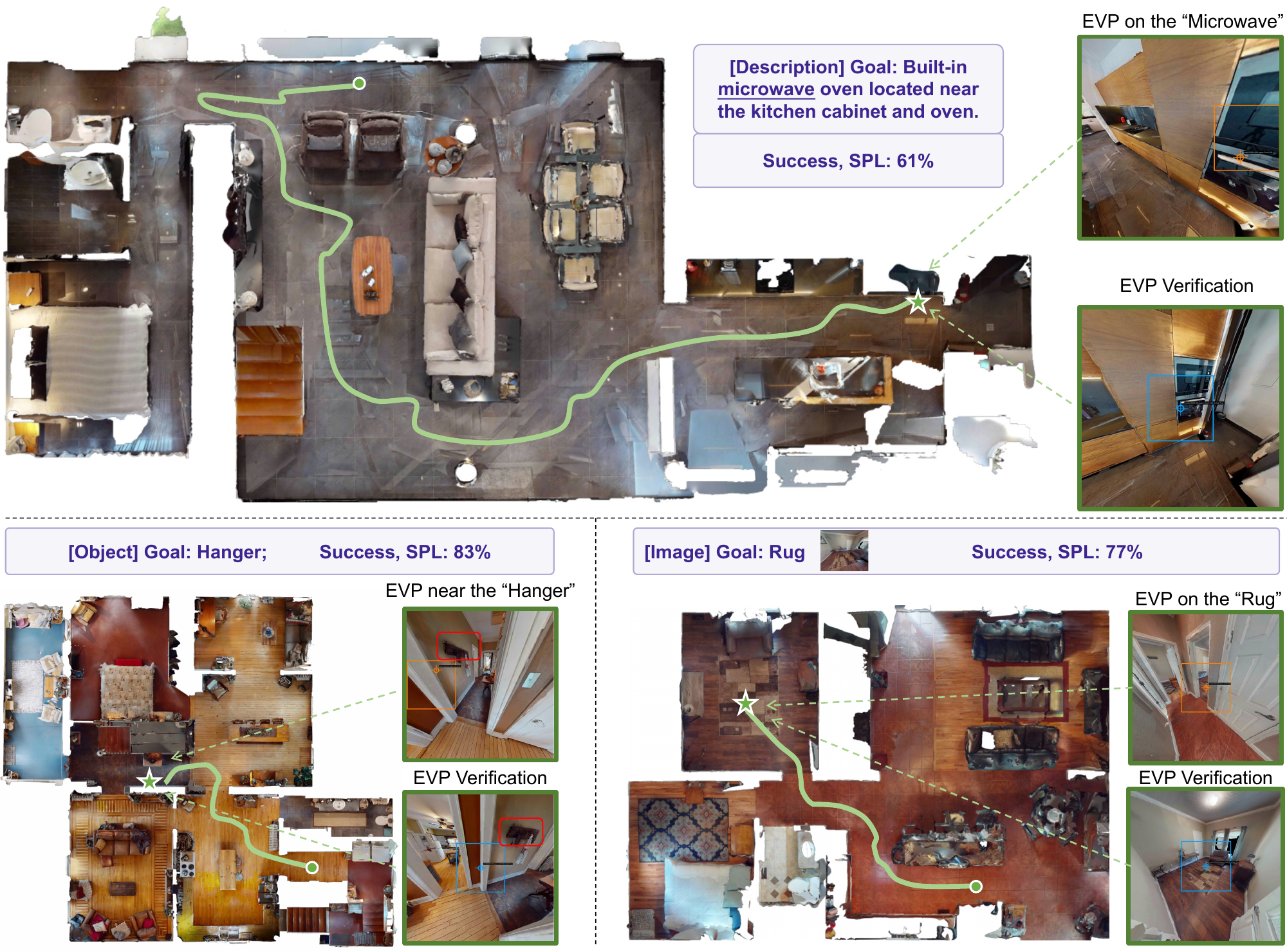}
    \caption{
    Qualitative examples of terminal EVP verification on GOAT-Bench. 
    The orange point and box denote the initially proposed, unverified EVP from the current observation, while the blue point and box denote the close-range verified EVP used for termination. 
    The red box in the bottom-left case marks the challenging hanger target.
    }
    \label{fig:exp_evp_prob}
\end{figure}

We inspect terminal-EVP cases in \cref{fig:exp_evp_prob} to illustrate how EVP complements object-level evidence when the target is absent from object memory. In these examples, the target objects are not detected, so \frkabbr cannot rely on a verified object proposal for commitment. Instead, the VLM first proposes an unverified EVP from the current view, marked by the orange box, and \frkabbr uses this point only as a provisional navigation cue. After moving toward the projected location, \frkabbr performs close-range verification and obtains the verified EVP, marked by the blue box, before terminating the task. Thus, EVP provides a detector-independent path from weak visual cues to locally verified spatial evidence.

The examples cover different goal specifications, including a description goal for the built-in microwave, an object goal for the hanger, and an image goal for the rug. In the microwave and rug cases, the initial EVP already indicates a task-relevant region that can guide the agent toward the target. The hanger case is more challenging: the true target is visually ambiguous, making it difficult to detect or recognize from the initial view. Although the EVP does not precisely hit the hanger, it is placed near the correct area and therefore still guides the agent to a useful close-range viewpoint for termination. These cases show that EVP can recover missed target cues by providing actionable point-level guidance.

\subsection{Analysis of Frontier Regulation}

\subsubsection{Quantitative Analysis}

To complement the task-level ablations in \cref{tab:ablation_aeqa_components}, we conduct behavior-level analyses on the A-EQA 41-subset to examine how AHS, VFR, and HLP shape frontier-selection behavior. For each mechanism, full \frkabbr is compared with the corresponding ablated variant.

\begin{figure}[ht]
\setlength{\belowcaptionskip}{-0.3cm}
    \centering
        \includegraphics[width=0.75\linewidth]{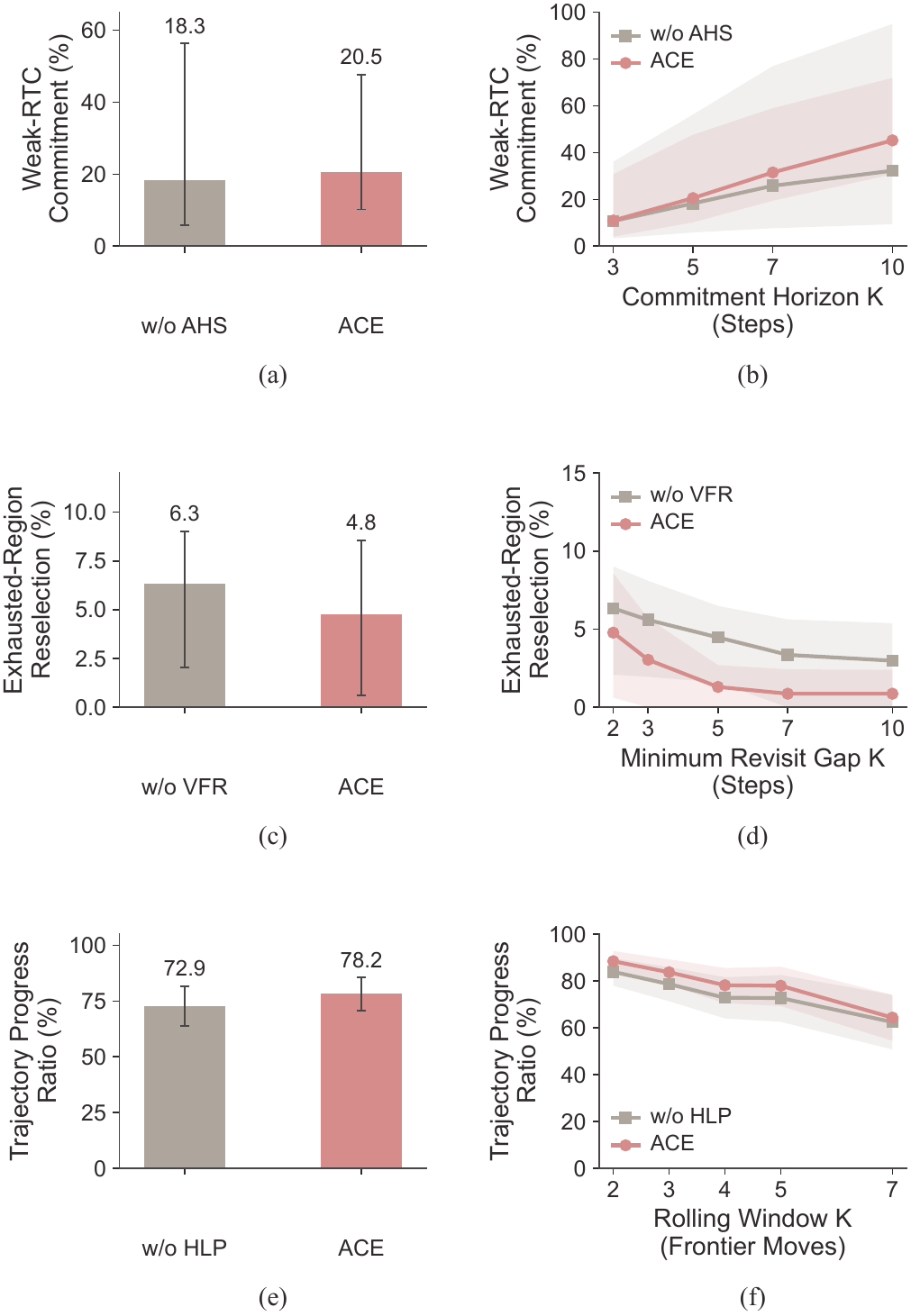}
    \caption{Behavior-level analyses.
    (a,b) Weak-RTC commitment at $K=5$ and varying commitment horizons.
    (c,d) Exhausted-region reselection at $K=2$ and varying minimum revisit gaps.
    (e,f) Trajectory progress ratio at $K=4$ and varying rolling frontier-move windows.
    Error bars and shaded bands denote 95\% episode-level bootstrap confidence intervals.}
    \label{fig:frontier_regulation_behavior}
\end{figure}

\textbf{AHS under Weak Reactive Cues.}
We collect frontier-selection decisions in which the selected frontier has a reactive task-cue (RTC) score below 0.5. For each decision, we examine the subsequent $K$ decision steps and regard it as successful if the agent reaches a snapshot containing sufficient answer evidence and commits within this horizon. The resulting weak-RTC commitment rate measures the proportion of such decisions that lead to evidence-supported commitment, using snapshot commitment as an observable indicator that answerable evidence has been reached. For the default setting of $K=5$, \frkabbr achieves a commitment rate of 20.5\%, compared with 18.3\% without AHS, as shown in \cref{fig:frontier_regulation_behavior}(a). The effect becomes more pronounced as the evaluation horizon increases. As shown in \cref{fig:frontier_regulation_behavior}(b), the two variants perform similarly at $K=3$ (11.0\% versus 10.8\%), whereas their rates diverge to 45.2\% and 32.3\%, respectively, at $K=10$. These results indicate that AHS primarily benefits subsequent evidence acquisition rather than immediate termination, by retaining frontier directions that remain promising despite weak currently visible task cues.

\subsubsection*{Exhausted-Region Reselection}
We define a frontier visit as low-yield if the observation acquired immediately afterward produces no candidate evidence. A subsequent frontier selection is identified as an exhausted-region reselection when it returns to within 8 map units of a previous low-yield visit, occurs at least $K$ frontier decisions later, and follows at least 8 map units of intervening travel. The travel-distance constraint excludes immediate local adjustments, while $K$ controls the temporal separation required for two visits to be regarded as repeated investment in the same region. The exhausted-region reselection rate is computed as the proportion of all frontier selections satisfying these conditions.

At the default setting of $K=2$, \frkabbr yields a reselection rate of 4.8\%, compared with 6.3\% for ``w/o VFR'', as shown in \cref{fig:frontier_regulation_behavior}(c). The difference becomes more pronounced when only longer-gap revisits are considered. In \cref{fig:frontier_regulation_behavior}(d), the reselection rates at $K=5$ are 1.3\% for \frkabbr and 4.5\% without VFR, while those at $K=10$ are 0.9\% and 3.0\%, respectively. This pattern indicates that VFR primarily suppresses persistent returns to previously inspected, low-yield regions rather than short-term local corrections. In addition, none of the frontier directions rejected by the visual frontier memory is subsequently recorded as productive in the evaluated trajectories, suggesting that the observed reduction in reselection is not accompanied by evident rejection of useful directions.

\subsubsection*{Local Trajectory Progress}
To characterize the effect of HLP on local trajectory progress, we construct all rolling windows $\Omega_e^K$ containing $K$ consecutive frontier moves in episode $e$ and evaluate each window using the displacement-to-path ratio defined in \cref{eq:progress_length}. To prevent episodes with longer trajectories from dominating the analysis, we first average the window-level ratios within each episode and then aggregate them across episodes:
\begin{equation}
    \mathrm{PR}(K)=
    \frac{1}{|\mathcal{E}_K|}
    \sum_{e\in\mathcal{E}_K}
    \frac{1}{|\Omega_e^K|}
    \sum_{\mathcal{W}\in\Omega_e^K}
    \frac{\Delta(\mathcal{W})}
    {L(\mathcal{W})+\epsilon},
    \label{eq:rolling_progress_ratio}
\end{equation}
where $\mathcal{E}_K$ contains the matched episodes for which both \frkabbr and the ``w/o HLP'' variant have at least one valid $K$-move window. Each episode contributes a single PR value, and all paired statistical analyses are performed at the episode level.

At the default setting of $K=4$, 16 matched episodes are available. As shown in \cref{fig:frontier_regulation_behavior}(e), \frkabbr increases the mean PR from 72.9\% to 78.2\%, corresponding to a paired mean improvement of 5.34 percentage points. \frkabbr achieves a higher PR in nine episodes, a lower PR in two episodes, and the same PR in five episodes. The improvement remains approximately five percentage points for $K=2$--$5$ in \cref{fig:frontier_regulation_behavior}(f), before narrowing at $K=7$, where fewer valid long-horizon windows are available. A two-sided paired Wilcoxon signed-rank test yields $W=13.0$ and $p=0.075$, with a matched-pairs rank-biserial correlation of $r_{\mathrm{rb}}=0.606$. Although the difference does not reach the conventional $0.05$ significance level, the effect magnitude and its predominantly positive episode-level direction indicate a consistent and practically meaningful improvement in short-horizon trajectory progress.

These analyses demonstrate the complementary effects of the three mechanisms. AHS preserves promising directions when visible task cues are weak. VFR suppresses excessive visits to exhausted regions. HLP promotes frontiers that achieve greater spatial progress per unit of traveled distance.

\subsubsection{Qualitative Analysis of AHS}

\begin{figure}[!ht]
    \centering
    \subfloat[A-EQA frontier selection.]{
        \includegraphics[width=1.0\linewidth]{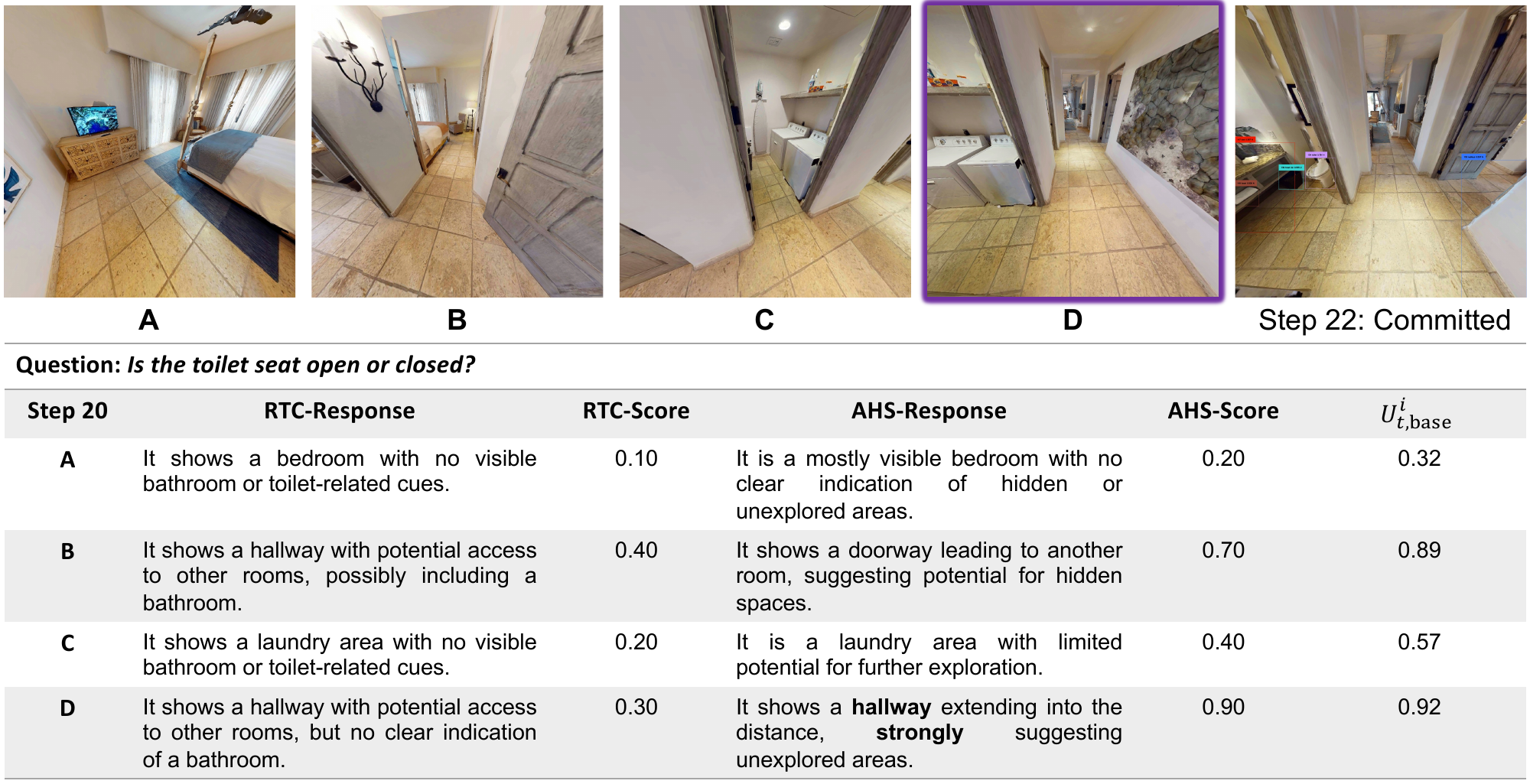}
        \label{fig:exp_pes_aeqa}}
    \vspace{0.2em}
    \subfloat[GOAT-Bench frontier selection.]{
        \includegraphics[width=1.0\linewidth]{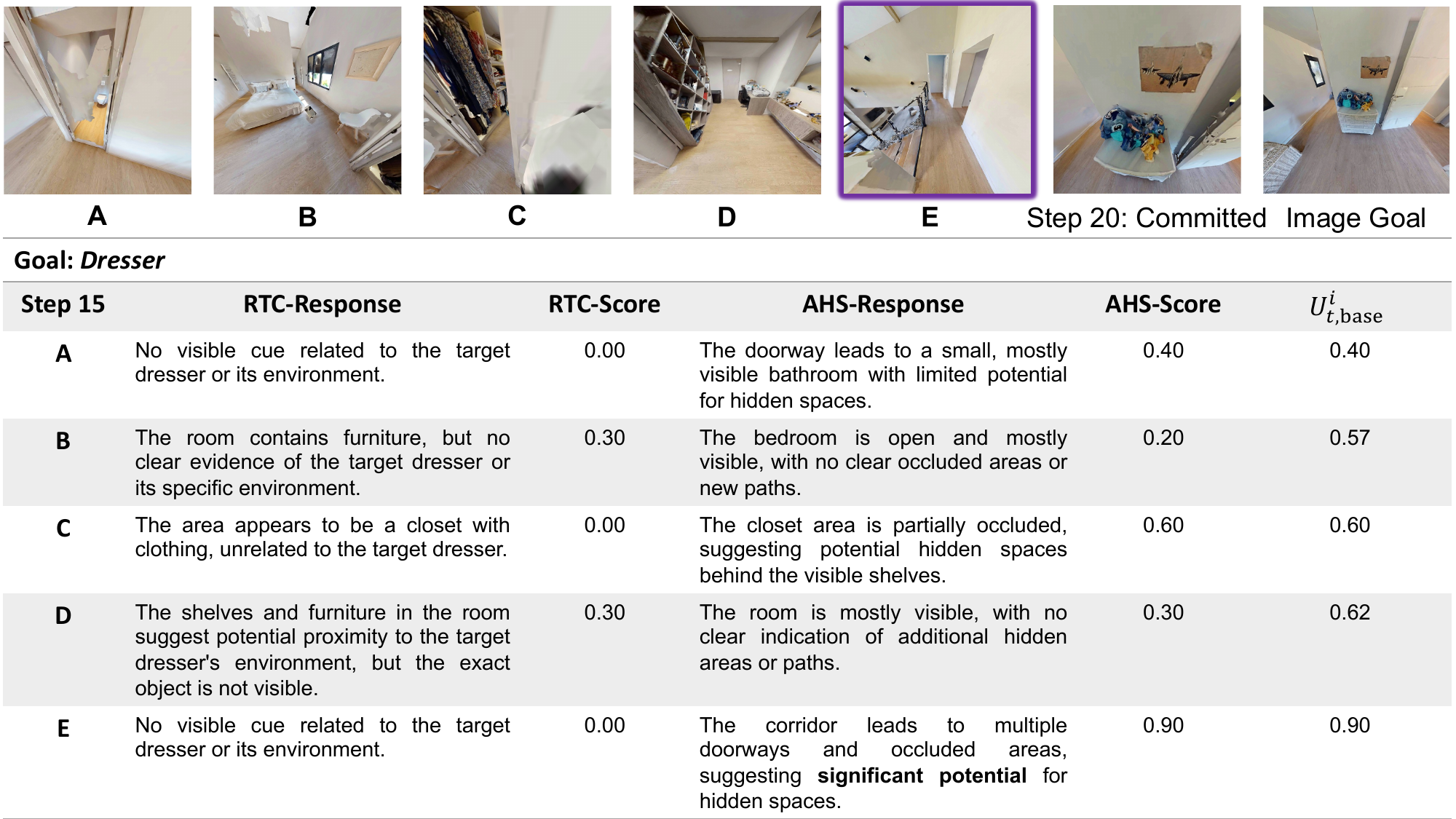}
        \label{fig:exp_pes_goat}}
    \caption{
    Qualitative examples of \cptwo. 
    Each case lists frontier candidates of a certain step with their RTC and AHS responses, numerical scores, and the resulting frontier utility. 
    The selected frontier is highlighted in purple. 
    }
    \label{fig:exp_pes}
\end{figure}

We provide qualitative examples in \cref{fig:exp_pes} to examine how the AHS score affects frontier selection. For each case, we list the frontier-facing candidate views, the VLM responses, numerical scores for the RTC score and the AHS score, and the resulting frontier utility. The selected frontier is highlighted in purple.

The two examples show a common pattern: when the current frontier-facing views contain weak or ambiguous task cues, the selected frontier is the one that provides stronger access to hidden space. In the A-EQA case (\cref{fig:exp_pes_aeqa}), the question concerns the toilet seat, but none of the current frontier views directly observe a toilet or bathroom. The RTC scores are therefore generally low or moderate. \frkabbr selects Frontier D because its view faces a hallway extending into the distance, which receives the highest AHS score and the highest final utility. After moving through this frontier, the agent soon reaches the observation used for answering. 

A similar behavior appears in the GOAT-Bench case (\cref{fig:exp_pes_goat}). The goal is to find an image-referred dresser, but no candidate frontier provides a clearly visible dresser cue, and several RTC scores are close to zero. \frkabbr selects Frontier E, not because it already shows the target, but because it leads to multiple doorways and occluded regions, giving it the strongest hidden-space potential. The subsequent exploration reaches the goal-related area and completes the task. These two cases indicate that, when direct visual evidence is absent, hidden-space potential can preserve a useful exploration route that would be missed by purely reactive semantic matching.

Overall, these examples show that \cptwo is not a greedy search over visible cues. The RTC score exploits what the frontier currently suggests, while the AHS score estimates what the frontier may reveal after being explored. When all visible cues are weak, hidden-space potential provides a critical tie-breaking and route-discovery signal, allowing \frkabbr to move toward frontier directions that are more likely to uncover task-relevant evidence.

\subsubsection{Qualitative Analysis of \cpthreeabbr}

\begin{figure}[ht]
\setlength{\belowcaptionskip}{-0.5cm}
    \centering
    \includegraphics[width=1.0\linewidth]{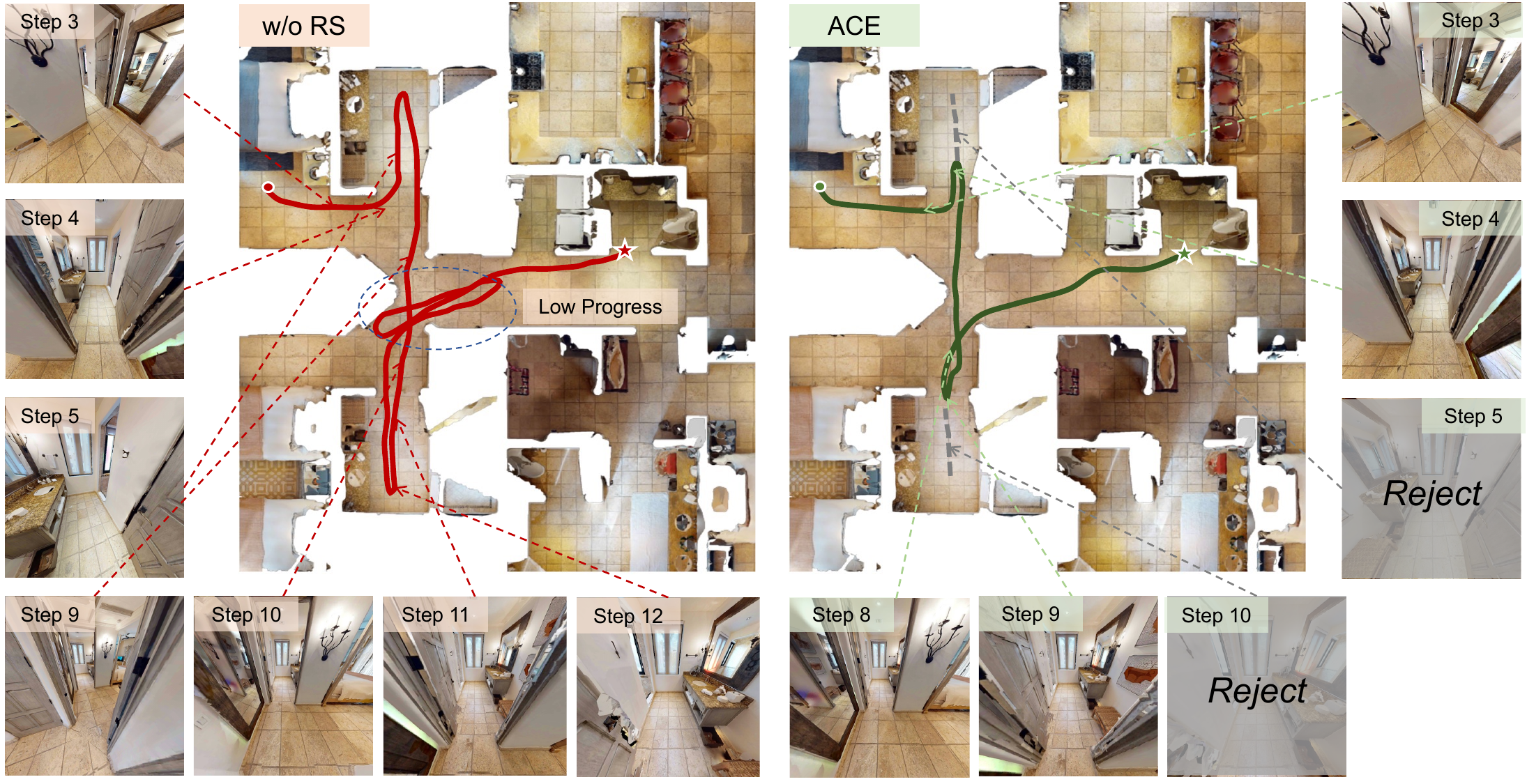}
    \caption{
    Comparison of the chosen frontier of each step between \textcolor{trajgreen}{\frkabbr} and \textcolor{red}{w/o \cpthreeabbr}.
    }
    \label{fig:exp_efr}
\end{figure}

We compare \frkabbr with a variant without  \cpthreeabbr to examine how frontier rejection changes exploration behavior. As shown in \cref{fig:exp_efr}, both agents start from a similar local region, but their behaviors diverge after the agent enters a restroom-like area where consecutive observations provide no task-relevant evidence. Without \cpthreeabbr, the agent continues to move deeper along the same local direction despite repeatedly observing visually similar and uninformative views. After this over-exploration, it further turns back and revisits nearby viewpoints, forming a low-progress trajectory around the same local area. Since no exhaustion feedback is accumulated, these frontier directions continue to consume navigation budget even though they have repeatedly failed to reveal useful evidence.

\frkabbr avoids this behavior by combining hard visual rejection with a soft history-aware penalty. Frontiers are rejected when their associated regions have been sufficiently inspected, repeatedly fail to provide task-relevant evidence, and belong to closed or weakly expandable contexts. Meanwhile, the history-aware penalty down-weights candidates that would extend recent low-displacement trajectories. Consequently, \frkabbr suppresses redundant deep inspection and subsequent revisits, and redirects exploration toward more productive continuations. This example shows that \cpthreeabbr helps \frkabbr decide when to move on from unproductive frontier directions, rather than repeatedly pursuing visually exhausted opportunities.

\subsection{Analysis of Successful and Failed Trajectories}

\begin{figure}[!ht]
\setlength{\belowcaptionskip}{-0.3cm}
    \centering
    \subfloat[Successful GOAT-Bench trajectories.]{
        \includegraphics[width=1.0\linewidth]{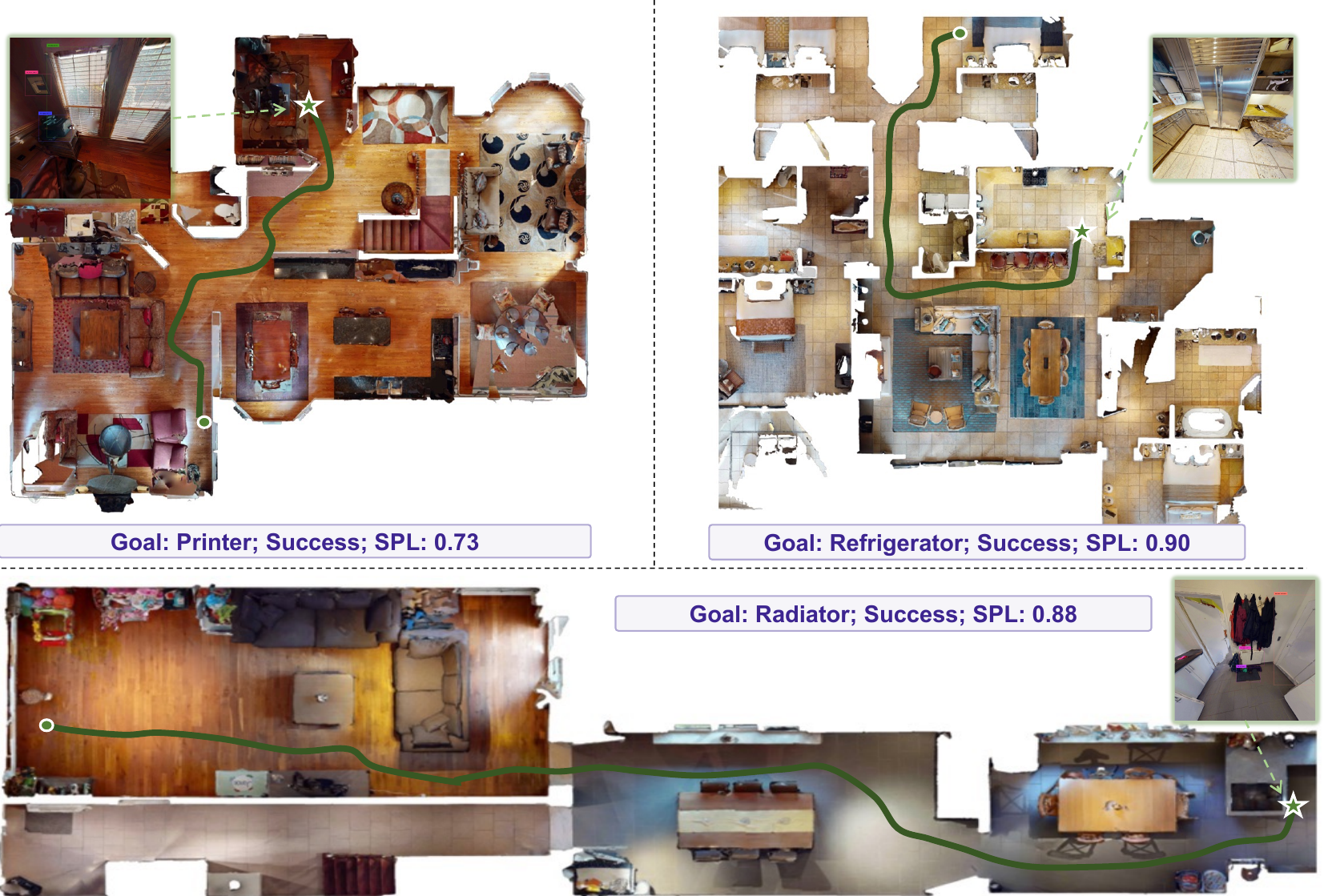}
        \label{fig:success_cases}}
    \vspace{0.1em}
    \subfloat[Failed GOAT-Bench trajectories.]{
        \includegraphics[width=1.0\linewidth]{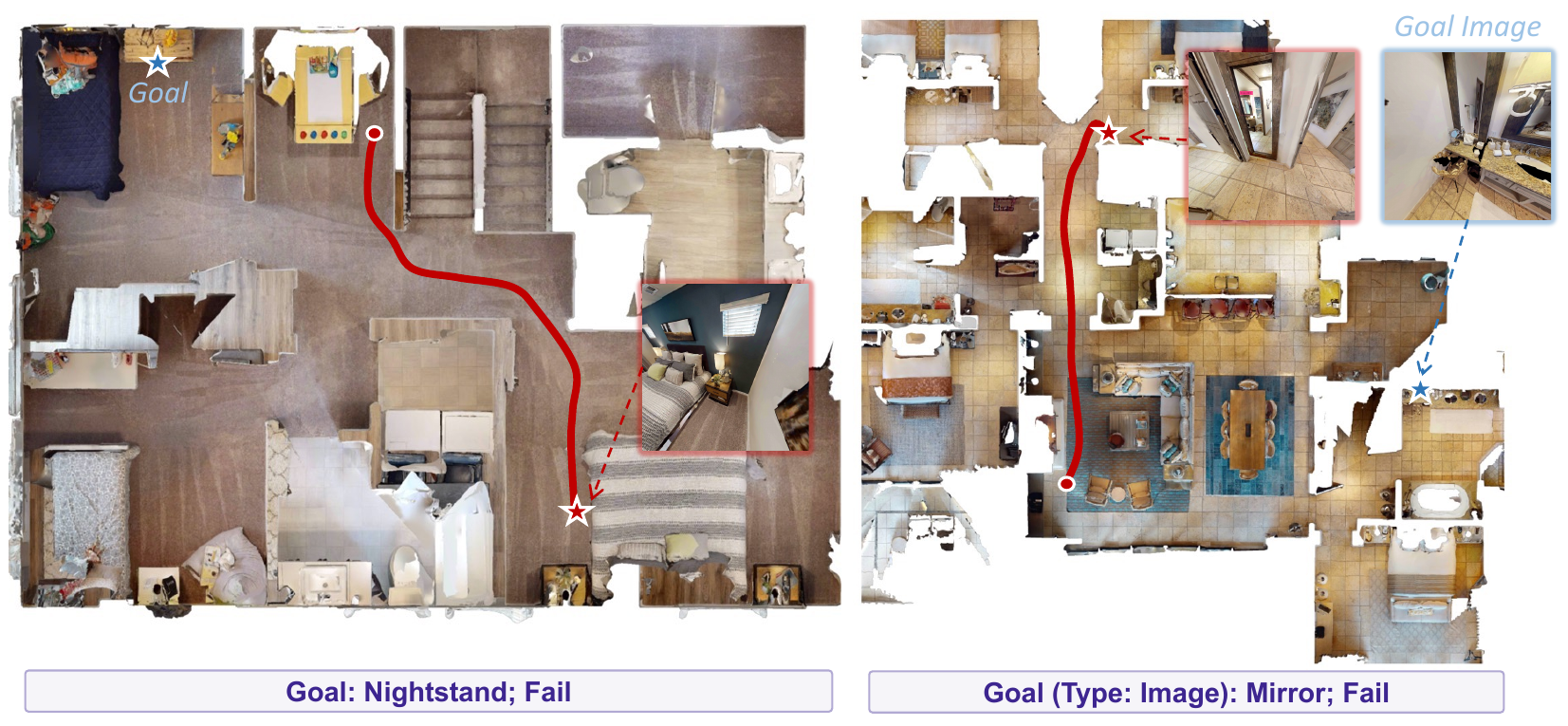}
        \label{fig:failure_cases}}
    \caption{
   Successful and failed trajectories on GOAT-Bench. 
    }
    \label{fig:success_failure_cases}
\end{figure}

We show representative successful and failed trajectories on GOAT-Bench in \cref{fig:success_failure_cases}. The bottom long-range radiator case in \cref{fig:success_cases} is particularly illustrative: the agent traverses multiple rooms and corridors before reaching the target-related area, showing that \frkabbr can support long-horizon target discovery rather than only local object matching.

The failure cases highlight three remaining challenges. First, cross-floor navigation is not explicitly handled. When the target requires going upstairs or downstairs, \frkabbr may continue searching on the current floor. This is also a common limitation of most compared methods, which mainly assume single-floor navigation. Second, incomplete annotations may lead to false negatives, as in the nightstand case of \cref{fig:failure_cases} where \frkabbr reaches a visually plausible target that is not annotated as valid. Third, image-goal navigation remains difficult when multiple instances share similar appearance and context. In the mirror case of \cref{fig:failure_cases}, \frkabbr stops near a visually similar mirror, but the benchmark target is another instance. These cases point to future improvements in floor-transition reasoning, annotation-aware analysis, and instance-level target disambiguation.

\section{Conclusion}
\label{sec:conclusion}

We present \frkabbr, a unified framework for reliable and efficient embodied exploration through spatially resolved decision making. \frkabbr couples \hpone with \hptwo to improve both termination and continuation. \cponefull establishes sufficient visual support through \cponeloc and \cponever across regional, object-level, and visual-point candidates, while \cptwofull and \cpthreefull preserve promising directions and suppress exhausted or low-progress ones by jointly considering prospective utility, accumulated visual coverage, and exploration feedback. Extensive experiments across active embodied question answering, single-goal navigation, and sequential multi-goal navigation demonstrate that \frkabbr consistently achieves strong task success and leading exploration efficiency among training-free methods. These results highlight spatially resolved decision making as a promising principle for effective and efficient embodied exploration.

{
    \small
    \bibliographystyle{IEEEtran}
    \bibliography{ace_ref}

@String(PAMI  = {IEEE Transactions on Pattern Analysis and Machine Intelligence})

@String(ICML  = {ICML})

@String(CVPR  = {CVPR})

@String(ICCV  = {ICCV})

@String(ECCV  = {ECCV})

@String(NIPS  = {NeurIPS})

@String(ICRA  = {ICRA})

@String(CORL  = {CoRL})

@String(IJRR  = {The International Journal of Robotics Research})

@String(ICLR  = {ICLR})

@String(AAAI = {AAAI})

@String(IROS = {IROS})

@inproceedings{armeni20193d,
  title={3d scene graph: A structure for unified semantics, 3d space, and camera},
  author={Armeni, Iro and He, Zhi-Yang and Gwak, JunYoung and Zamir, Amir R and Fischer, Martin and Malik, Jitendra and Savarese, Silvio},
  booktitle=CVPR,
  year={2019}
}

@article{rosinol2021kimera,
  title={Kimera: From SLAM to spatial perception with 3D dynamic scene graphs},
  author={Rosinol, Antoni and Violette, Andrew and Abate, Marcus and Hughes, Nathan and Chang, Yun and Shi, Jingnan and Gupta, Arjun and Carlone, Luca},
  journal=IJRR,
  volume={40},
  number={12-14},
  pages={1510--1546},
  year={2021}
}

@inproceedings{liu2023bird,
  title={Bird's-eye-view scene graph for vision-language navigation},
  author={Liu, Rui and Wang, Xiaohan and Wang, Wenguan and Yang, Yi},
  booktitle=ICCV,
  year={2023}
}

@inproceedings{yin2024sg,
  title={Sg-nav: Online 3d scene graph prompting for llm-based zero-shot object navigation},
  author={Yin, Hang and Xu, Xiuwei and Wu, Zhenyu and Zhou, Jie and Lu, Jiwen},
  booktitle=NIPS,
  year={2024}
}

@InProceedings{saxena2024grapheqa,
  title = 	 {GraphEQA: Using 3D Semantic Scene Graphs for Real-time Embodied Question Answering},
  author =       {Saxena, Saumya and Buchanan, Blake and Paxton, Chris and Liu, Peiqi and Chen, Bingqing and Vaskevicius, Narunas and Palmieri, Luigi and Francis, Jonathan and Kroemer, Oliver},
  booktitle = CoRL,
  year = 	 {2025},
}

@inproceedings{werby2024hovsg,
  title={Hierarchical Open-Vocabulary 3D Scene Graphs for Language-Grounded Robot Navigation},
  author={Werby, Abdelrhman and Huang, Chenguang and Buchner, Martin and Valada, Abhinav and Burgard, Wolfram},
  booktitle={ICRA Workshop},
  year={2024}
}

@inproceedings{wang2024karma,
  title={Karma: Augmenting embodied ai agents with long-and-short term memory systems},
  author={Wang, Zixuan and Yu, Bo and Zhao, Junzhe and Sun, Wenhao and Hou, Sai and Liang, Shuai and Hu, Xing and Han, Yinhe and Gan, Yiming},
  booktitle=ICRA,
  year={2025}
}

@inproceedings{shafiullah2022clip,
  title={Clip-fields: Weakly supervised semantic fields for robotic memory},
  author={Shafiullah, Nur Muhammad Mahi and Paxton, Chris and Pinto, Lerrel and Chintala, Soumith and Szlam, Arthur},
  booktitle={ICRA Workshop},
  year={2023}
}

@inproceedings{yamazaki2024open,
  title={Open-fusion: Real-time open-vocabulary 3d mapping and queryable scene representation},
  author={Yamazaki, Kashu and Hanyu, Taisei and Vo, Khoa and Pham, Thang and Tran, Minh and Doretto, Gianfranco and Nguyen, Anh and Le, Ngan},
  booktitle=ICRA,
  year={2024}
}

@article{jatavallabhula2023conceptfusion,
  title={Conceptfusion: Open-set multimodal 3d mapping},
  author={Jatavallabhula, Krishna Murthy and Kuwajerwala, Alihusein and Gu, Qiao and Omama, Mohd and Chen, Tao and Maalouf, Alaa and Li, Shuang and Iyer, Ganesh and Saryazdi, Soroush and Keetha, Nikhil and others},
  journal={arXiv:2302.07241},
  year={2023}
}

@inproceedings{kerr2023lerf,
  title={Lerf: Language embedded radiance fields},
  author={Kerr, Justin and Kim, Chung Min and Goldberg, Ken and Kanazawa, Angjoo and Tancik, Matthew},
  booktitle=ICCV,
  year={2023}
}

@article{lei2025gaussnav,
  title={Gaussnav: Gaussian splatting for visual navigation},
  author={Lei, Xiaohan and Wang, Min and Zhou, Wengang and Li, Houqiang},
  journal=PAMI,
  volume={47},
  number={5},
  pages={4108--4121},
  year={2025}
}

@article{huang2025multimodal,
  title={Multimodal spatial language maps for robot navigation and manipulation},
  author={Huang, Chenguang and Mees, Oier and Zeng, Andy and Burgard, Wolfram},
  journal=IJRR,
  year={2025}
}

@inproceedings{huang23vlmaps,
    title={Visual Language Maps for Robot Navigation},
    author={Huang, Chenguang and Mees, Oier and Zeng, Andy and Burgard, Wolfram},
    booktitle = ICRA,
    year={2023}
}

@inproceedings{yang20253d,
  title={3D-mem: 3D scene memory for embodied exploration and reasoning},
  author={Yang, Yuncong and Yang, Han and Zhou, Jiachen and Chen, Peihao and Zhang, Hongxin and Du, Yilun and Gan, Chuang},
  booktitle=CVPR,
  year={2025}
}

@inproceedings{gu2024conceptgraphs,
  title={Conceptgraphs: Open-vocabulary 3d scene graphs for perception and planning},
  author={Gu, Qiao and Kuwajerwala, Ali and Morin, Sacha and Jatavallabhula, Krishna Murthy and Sen, Bipasha and Agarwal, Aditya and Rivera, Corban and Paul, William and Ellis, Kirsty and Chellappa, Rama and others},
  booktitle=ICRA,
  year={2024}
}

@inproceedings{jiang2024roboexp,
    title={RoboEXP: Action-Conditioned Scene Graph via Interactive Exploration for Robotic Manipulation},
    author={Jiang, Hanxiao and Huang, Binghao and Wu, Ruihai and Li, Zhuoran and Garg, Shubham and Nayyeri, Hooshang and Wang, Shenlong and Li, Yunzhu},
    booktitle=CORL,
    year={2024}
}

@inproceedings{majumdar2024openeqa,
  title={Openeqa: Embodied question answering in the era of foundation models},
  author={Majumdar, Arjun and Ajay, Anurag and Zhang, Xiaohan and Putta, Pranav and Yenamandra, Sriram and Henaff, Mikael and Silwal, Sneha and Mcvay, Paul and Maksymets, Oleksandr and Arnaud, Sergio and others},
  booktitle=CVPR,
  year={2024}
}

@article{ren2024explore,
  title={Explore until confident: Efficient exploration for embodied question answering},
  author={Ren, Allen Z and Clark, Jaden and Dixit, Anushri and Itkina, Masha and Majumdar, Anirudha and Sadigh, Dorsa},
  journal={arXiv:2403.15941},
  year={2024}
}

@inproceedings{das2018embodiedqa,
  title={Embodied Question Answering},
  author={Das, Abhishek and Datta, Samyak and Gkioxari, Georgia and Lee, Stefan and Parikh, Devi and Batra, Dhruv},
  booktitle=CVPR,
  year={2018}
}

@article{batra2020objectnav,
  title={ObjectNav Revisited: On Evaluation of Embodied Agents Navigating to Objects},
  author={Batra, Dhruv and Gokaslan, Aaron and Kembhavi, Aniruddha and Maksymets, Oleksandr and Mottaghi, Roozbeh and Savva, Manolis and Toshev, Alexander and Wijmans, Erik},
  journal={arXiv:2006.13171},
  year={2020}
}

@article{krantz2022instanceimagenav,
  title={Instance-Specific Image Goal Navigation: Training Embodied Agents to Find Object Instances},
  author={Krantz, Jacob and Lee, Stefan and Malik, Jitendra and Batra, Dhruv and Chaplot, Devendra Singh},
  journal={arXiv:2211.15876},
  year={2022}
}

@inproceedings{anderson2018vision,
  title={Vision-and-Language Navigation: Interpreting Visually-Grounded Navigation Instructions in Real Environments},
  author={Anderson, Peter and Wu, Qi and Teney, Damien and Bruce, Jake and Johnson, Mark and Sunderhauf, Niko and Reid, Ian and Gould, Stephen and van den Hengel, Anton},
  booktitle=CVPR,
  year={2018}
}

@article{wani2020multion,
  title={MultiON: Benchmarking Semantic Map Memory Using Multi-Object Navigation},
  author={Wani, Saim and Patel, Shivansh and Jain, Unnat and Chang, Angel X and Savva, Manolis},
  journal={arXiv:2012.03912},
  year={2020}
}

@inproceedings{cheng2024efficienteqa,
  title={Efficienteqa: An efficient approach to open-vocabulary embodied question answering},
  author={Cheng, Kai and Li, Zhengyuan and Sun, Xingpeng and Min, Byung-Cheol and Bedi, Amrit Singh and Bera, Aniket},
  booktitle=IROS,
  year={2025}
}

@inproceedings{yokoyama2023vlfm,
  title={Vlfm: Vision-language frontier maps for zero-shot semantic navigation},
  author={Yokoyama, Naoki and Ha, Sehoon and Batra, Dhruv and Wang, Jiuguang and Bucher, Bernadette},
  booktitle=ICRA,
  year={2024}
}

@inproceedings{cui2024frontier,
  title={Frontier-Enhanced Topological Memory with Improved Exploration Awareness for Embodied Visual Navigation},
  author={Cui, Xinru and Liu, Qiming and Liu, Zhe and Wang, Hesheng},  
  booktitle=ECCV,
  year={2024}
}

@inproceedings{gadre2022cow,
  title={Cows on pasture: Baselines and benchmarks for language-driven zero-shot object navigation},
  author={Gadre, Samir Yitzhak and Wortsman, Mitchell and Ilharco, Gabriel and Schmidt, Ludwig and Song, Shuran},
  booktitle=CVPR,
  year={2023}
}

@inproceedings{shah2022lmnav,
  title={LM-Nav: Robotic Navigation with Large Pre-Trained Models of Language, Vision, and Action},
  author={Shah, Dhruv and Osinski, Blazej and Ichter, Brian and Levine, Sergey},
  booktitle=CoRL,
  year={2023}
}

@inproceedings{ziliotto2025tango,
  title={Tango: training-free embodied ai agents for open-world tasks},
  author={Ziliotto, Filippo and Campari, Tommaso and Serafini, Luciano and Ballan, Lamberto},
  booktitle=CVPR,
  year={2025}
}

@article{zhai2025memoryeqa,
  title={Memory-Centric Embodied Question Answer},
  author={Zhai, Mingliang and Gao, Zhi and Wu, Yuwei and Jia, Yunde},
  journal={arXiv:2505.13948},
  year={2025}
}

@inproceedings{fan2025embodiedvideoagent,
  title={Embodied videoagent: Persistent memory from egocentric videos and embodied sensors enables dynamic scene understanding},
  author={Fan, Yue and Ma, Xiaojian and Su, Rongpeng and Guo, Jun and Wu, Rujie and Chen, Xi and Li, Qing},
  booktitle=ICCV,
  year={2025}
}

@article{lu2026gsmem,
  title={GSMem: 3D Gaussian Splatting as Persistent Spatial Memory for Zero-Shot Embodied Exploration and Reasoning},
  author={Lu, Yiren and Du, Yi and Liu, Disheng and Zhou, Yunlai and Wang, Chen and Yin, Yu},
  journal={arXiv:2603.19137},
  year={2026}
}

@article{chen2026hgr,
  title={Hypothesis Graph Refinement: Hypothesis-Driven Exploration with Cascade Error Correction for Embodied Navigation},
  author={Chen, Peixin and Zhang, Guoxi and Ma, Jianwei and Li, Qing},
  journal={arXiv:2604.04108},
  year={2026}
}

@inproceedings{hu2026astranav,
  title={AstraNav-Memory: Contexts Compression for Long Memory},
  author={Hu, Junjun and Xue, Xinda and Ren, Botao and Luo, Minghua and Chen, Jintao and Bai, Haochen and You, Liangliang and Xu, Mu},
  booktitle=CVPR,
  year={2026}
}

@inproceedings{yang2025mindjourney,
  title={Mindjourney: Test-time scaling with world models for spatial reasoning},
  author={Yang, Yuncong and Liu, Jiageng and Zhang, Zheyuan and Zhou, Siyuan and Tan, Reuben and Yang, Jianwei and Du, Yilun and Gan, Chuang},
  booktitle=NIPS,
  year={2026}
}

@inproceedings{zhu2025mtu,
  title = {Move to Understand a 3D Scene: Bridging Visual Grounding and Exploration for Efficient and Versatile Embodied Navigation},
  author = {Zhu, Ziyu and Wang, Xilin and Li, Yixuan and Zhang, Zhuofan and Ma, Xiaojian and Chen, Yixin and Jia, Baoxiong and Liang, Wei and Yu, Qian and Deng, Zhidong and Huang, Siyuan and Li, Qing},
  booktitle = ICCV,
  year = {2025}  
}

@inproceedings{ginting2025enter,
  title={Enter the Mind Palace: Reasoning and Planning for Long-term Active Embodied Question Answering},
  author={Ginting, Muhammad Fadhil and Kim, Dong-Ki and Meng, Xiangyun and Reinke, Andrzej Marek and Krishna, Bandi Jai and Kayhani, Navid and Peltzer, Oriana and Fan, David and Shaban, Amirreza and Kim, Sung-Kyun and others},
  booktitle=CoRL,
  year={2025}
}

@article{ruan2026brain,
  title={Brain-inspired spatial intelligence for embodied agents},
  author={Ruan, Shouwei and Wang, Liyuan and Kang, Caixin and Zhu, Qihui and Liu, Songming and Wei, Xingxing and Su, Hang},
  journal={Nature Communications},
  year={2026},
}

@inproceedings{yin2025unigoal,
  title={Unigoal: Towards universal zero-shot goal-oriented navigation},
  author={Yin, Hang and Xu, Xiuwei and Zhao, Linqing and Wang, Ziwei and Zhou, Jie and Lu, Jiwen},
  booktitle=CVPR,
  year={2025}
}

@inproceedings{zhang2025embodied,
  title={Embodied navigation foundation model},
  author={Zhang, Jiazhao and Li, Anqi and Qi, Yunpeng and Li, Minghan and Liu, Jiahang and Wang, Shaoan and Liu, Haoran and Zhou, Gengze and Wu, Yuze and Li, Xingxing and others},
  booktitle=ICLR,
  year={2026}
}

@article{zhang2025hoz++,
  title={HOZ++: Versatile Hierarchical Object-to-Zone Graph for Object Navigation},
  author={Zhang, Sixian and Song, Xinhang and Yu, Xinyao and Bai, Yubing and Guo, Xinlong and Li, Weijie and Jiang, Shuqiang},
  journal=PAMI,
  volume={47},
  number={7},
  pages={5958-5975},
  year={2025},
  publisher={IEEE}
}

@inproceedings{khanna2024goat,
  title={Goat-bench: A benchmark for multi-modal lifelong navigation},
  author={Khanna, Mukul and Ramrakhya, Ram and Chhablani, Gunjan and Yenamandra, Sriram and Gervet, Theophile and Chang, Matthew and Kira, Zsolt and Chaplot, Devendra Singh and Batra, Dhruv and Mottaghi, Roozbeh},
  booktitle=CVPR,
  year={2024}
}

@inproceedings{song2025towards,
  title={Towards long-horizon vision-language navigation: Platform, benchmark and method},
  author={Song, Xinshuai and Chen, Weixing and Liu, Yang and Chen, Weikai and Li, Guanbin and Lin, Liang},
  booktitle=CVPR,
  year={2025}
}

@inproceedings{wu2025spatial,
  title={Spatial-mllm: Boosting mllm capabilities in visual-based spatial intelligence},
  author={Wu, Diankun and Liu, Fangfu and Hung, Yi-Hsin and Duan, Yueqi},
  booktitle=NIPS,
  year={2026}
}

@article{bai2025qwen2,
  title={Qwen2.5-vl technical report},
  author={Bai, Shuai and Chen, Keqin and Liu, Xuejing and Wang, Jialin and Ge, Wenbin and Song, Sibo and Dang, Kai and Wang, Peng and Wang, Shijie and Tang, Jun and others},
  journal={arXiv:2502.13923},
  year={2025}
}

@article{achiam2023gpt,
  title={Gpt-4 technical report},
  author={OpenAI},
  journal={arXiv:2303.08774},
  year={2023}
}

@article{li2024llava,
  title={Llava-onevision: Easy visual task transfer},
  author={Li, Bo and Zhang, Yuanhan and Guo, Dong and Zhang, Renrui and Li, Feng and Zhang, Hao and Zhang, Kaichen and Zhang, Peiyuan and Li, Yanwei and Liu, Ziwei and others},
  journal={arXiv:2408.03326},
  year={2024}
}

@article{team2024gemini,
  title={Gemini 1.5: Unlocking multimodal understanding across millions of tokens of context},
  author={Team, Gemini and Georgiev, Petko and Lei, Ving Ian and Burnell, Ryan and Bai, Libin and Gulati, Anmol and Tanzer, Garrett and Vincent, Damien and Pan, Zhufeng and Wang, Shibo and others},
  journal={arXiv:2403.05530},
  year={2024}
}

@article{gottlieb2018active_sampling,
  title={Towards a neuroscience of active sampling and curiosity},
  author={Gottlieb, Jacqueline and Oudeyer, Pierre-Yves},
  journal={Nature Reviews Neuroscience},
  volume={19},
  number={12},
  pages={758--770},
  year={2018},
  doi={10.1038/s41583-018-0078-0}
}

@article{petitet2021active_sampling,
  title={The computational cost of active information sampling before decision-making under uncertainty},
  author={Petitet, Pierre and Attaallah, Bahaaeddin and Manohar, Sanjay G. and others},
  journal={Nature Human Behaviour},
  volume={5},
  number={7},
  pages={935--946},
  year={2021},
  doi={10.1038/s41562-021-01116-6}
}

@article{corbetta2002control,
  title={Control of goal-directed and stimulus-driven attention in the brain},
  author={Corbetta, Maurizio and Shulman, Gordon L.},
  journal={Nature Reviews Neuroscience},
  volume={3},
  number={3},
  pages={201--215},
  year={2002},
  doi={10.1038/nrn755}
}

@article{gottlieb2013information,
  title={Information-seeking, curiosity, and attention: computational and neural mechanisms},
  author={Gottlieb, Jacqueline and Oudeyer, Pierre-Yves and Lopes, Manuel and Baranes, Adrien},
  journal={Trends in Cognitive Sciences},
  volume={17},
  number={11},
  pages={585--593},
  year={2013},
  doi={10.1016/j.tics.2013.09.001}
}

@article{balsdon2020confidence,
  title={Confidence controls perceptual evidence accumulation},
  author={Balsdon, Tarryn and Wyart, Valentin and Mamassian, Pascal},
  journal={Nature Communications},
  volume={11},
  number={1},
  pages={1753},
  year={2020},
  doi={10.1038/s41467-020-15561-w}
}

@article{bajcsy2018revisiting,
  title={Revisiting Active Perception},
  author={Bajcsy, Ruzena and Aloimonos, Yiannis and Tsotsos, John K},
  journal={Autonomous Robots},
  volume={42},
  number={2},
  pages={177--196},
  year={2018}
}

@inproceedings{pathak2017curiosity,
  title={Curiosity-driven Exploration by Self-supervised Prediction},
  author={Pathak, Deepak and Agrawal, Pulkit and Efros, Alexei A and Darrell, Trevor},
  booktitle=ICML,
  year={2017}
}

@article{metanav2026,
  title={Stop Wandering: Efficient Vision-Language Navigation via Metacognitive Reasoning},
  author={Li, Xueying and Lyu, Feng and Wu, Hao and Liu, Mingliu and Liu, Jia-Nan and Liu, Guozi},
  journal={arXiv:2604.02318},
  year={2026}
}

@article{zhang2025reexplore,
  title={ReEXplore: Improving MLLMs for Embodied Exploration with Contextualized Retrospective Experience Replay},
  author={Zhang, Gengyuan and Ding, Mingcong and Wu, Jingpei and Liao, Routong and Tresp, Volker},
  journal={arXiv:2511.19033},
  year={2025}
}

@article{li2026himm,
  title={HIMM: Human-Inspired Long-Term Memory Modeling for Embodied Exploration and Question Answering},
  author={Li, Ji and Wang, Bo and Xia, Jing and Li, Mingyi and Hu, Shiyan},
  journal={arXiv:2602.15513},
  year={2026}
}

@inproceedings{wang2026scope,
  title={Expand Your SCOPE: Semantic Cognition over Potential-Based Exploration for Embodied Visual Navigation},
  author={Wang, Ningnan and Chen, Weihuang and Chen, Liming and Ji, Haoxuan and Guo, Zhongyu and Zhang, Xuchong and Sun, Hongbin},
  booktitle=AAAI,
  year={2026}
}

@inproceedings{ehsani2024spoc,
  title={SPOC: Imitating Shortest Paths in Simulation Enables Effective Navigation and Manipulation in the Real World},
  author={Ehsani, Kiana and Gupta, Tanmay and Hendrix, Rose and Salvador, Jordi and Weihs, Luca and Zeng, Kuo-Hao and Singh, Kunal Pratap and Kim, Yejin and Han, Winson and Herrasti, Alvaro and others},
  booktitle=CVPR,
  year={2024}
}

@article{pistone2024modelling,
  title={Modelling and control of manipulators for inspection and maintenance in challenging environments: A literature review},
  author={Pistone, Alessandro and Ludovico, Daniele and De Mari Casareto Dal Verme, Lorenzo and Leggieri, Sergio and Canali, Carlo and Caldwell, Darwin G.},
  journal={Annual Reviews in Control},
  volume={57},
  pages={100949},
  year={2024},
  doi={10.1016/j.arcontrol.2024.100949}
}

@inproceedings{dai2026history,
  title={History to Future: Evolving Agent with Experience and Thought for Zero-shot Vision-and-Language Navigation},
  author={Dai, Guangzhao and Wang, Shuo and Wang, Zihan and Xie, Guo-Sen and Yang, Yang and Pan, Jinshan and Sun, Qianru and Shu, Xiangbo},
  booktitle=CVPR,
  year={2026}
}

@article{wang2026imaginenav++,
  title={ImagineNav++: Prompting Vision-Language Models as Embodied Navigator through Scene Imagination},
  author={Wang, Teng and Zhao, Xinxin and Cai, Wenzhe and Sun, Changyin},
  journal=PAMI,
  volume={48},
  number={9},
  pages={11166--11182},
  year={2026},
  publisher={IEEE}
}

@article{hu2025imaginative,
  title={Imaginative world modeling with scene graphs for embodied agent navigation},
  author={Hu, Yue and Wu, Junzhe and Xu, Ruihan and Liu, Hang and Xi, Avery and Liu, Henry X and Vasudevan, Ram and Ghaffari, Maani},
  journal={arXiv preprint arXiv:2508.06990},
  year={2025}
}

@inproceedings{yokoyama2024hm3d,
  title={Hm3d-ovon: A dataset and benchmark for open-vocabulary object goal navigation},
  author={Yokoyama, Naoki and Ramrakhya, Ram and Das, Abhishek and Batra, Dhruv and Ha, Sehoon},
  booktitle=IROS,
  year={2024}
}

@article{zhou2025reinforced,
  title={Reinforced mllm: A survey on rl-based reasoning in multimodal large language models},
  author={Zhou, Guanghao and Qiu, Panjia and Chen, Cen and Wang, Jie and Yang, Zheming and Xu, Jian and Qiu, Minghui},
  journal={arXiv preprint arXiv:2504.21277},
  year={2025}
}

@inproceedings{wijmans2019embodied,
  title={Embodied question answering in photorealistic environments with point cloud perception},
  author={Wijmans, Erik and Datta, Samyak and Maksymets, Oleksandr and Das, Abhishek and Gkioxari, Georgia and Lee, Stefan and Essa, Irfan and Parikh, Devi and Batra, Dhruv},
  booktitle=CVPR,
  year={2019}
}

@inproceedings{pomerleau1988alvinn,
  title={Alvinn: An autonomous land vehicle in a neural network},
  author={Pomerleau, Dean A},
  booktitle=NIPS,
  year={1988}
}

@inproceedings{ross2011reduction,
  title={A reduction of imitation learning and structured prediction to no-regret online learning},
  author={Ross, St{\'e}phane and Gordon, Geoffrey and Bagnell, Drew},
  booktitle={AISTATS},
  year={2011},
}

@article{zhang2024uni,
  title={Uni-navid: A video-based vision-language-action model for unifying embodied navigation tasks},
  author={Zhang, Jiazhao and Wang, Kunyu and Wang, Shaoan and Li, Minghan and Liu, Haoran and Wei, Songlin and Wang, Zhongyuan and Zhang, Zhizheng and Wang, He},
  journal={arXiv preprint arXiv:2412.06224},
  year={2024}
}

@inproceedings{zemskova2026ovsegdt,
  title={OVSegDT: Segmenting Transformer for Open-Vocabulary Object Goal Navigation},
  author={Zemskova, Tatiana and Staroverov, Aleksei and Yudin, Dmitry and Panov, Aleksandr},
  booktitle=CVPR,
  year={2026}
}

@inproceedings{wang2026dynam3d,
  title={Dynam3d: Dynamic layered 3d tokens empower vlm for vision-and-language navigation},
  author={Wang, Zihan and Lee, Seungjun and Lee, Gim Hee},
  booktitle=NIPS,
  year={2026}
}

@inproceedings{savva2019habitat,
  title={Habitat: A platform for embodied ai research},
  author={Savva, Manolis and Kadian, Abhishek and Maksymets, Oleksandr and Zhao, Yili and Wijmans, Erik and Jain, Bhavana and Straub, Julian and Liu, Jia and Koltun, Vladlen and Malik, Jitendra and others},
  booktitle=ICCV,
  year={2019}
}

@inproceedings{cheng2024yolo,
  title={Yolo-world: Real-time open-vocabulary object detection},
  author={Cheng, Tianheng and Song, Lin and Ge, Yixiao and Liu, Wenyu and Wang, Xinggang and Shan, Ying},
  booktitle=CVPR,
  year={2024}
}

@inproceedings{kirillov2023segment,
  title={Segment anything},
  author={Kirillov, Alexander and Mintun, Eric and Ravi, Nikhila and Mao, Hanzi and Rolland, Chloe and Gustafson, Laura and Xiao, Tete and Whitehead, Spencer and Berg, Alexander C and Lo, Wan-Yen and others},
  booktitle=ICCV,
  year={2023}
}

@inproceedings{radford2021learning,
  title={Learning transferable visual models from natural language supervision},
  author={Radford, Alec and Kim, Jong Wook and Hallacy, Chris and Ramesh, Aditya and Goh, Gabriel and Agarwal, Sandhini and Sastry, Girish and Askell, Amanda and Mishkin, Pamela and Clark, Jack and others},
  booktitle=ICML,
  year={2021}
}

@article{team2026qwen3,
  title={Qwen3.5-omni technical report},
  author={Team, Qwen},
  journal={arXiv preprint arXiv:2604.15804},
  year={2026}
}
}



\end{document}